\documentclass[10pt,twocolumn,letterpaper]{article}

\usepackage{wacv}
\usepackage{times}
\usepackage{epsfig}
\usepackage{graphicx}
\usepackage{amsmath}
\usepackage{amssymb}
\usepackage{booktabs}
\usepackage{tabularx}
\usepackage{multirow}
\usepackage{threeparttable}
\usepackage{dsfont}
\usepackage{gensymb}
\usepackage[page]{appendix}
\usepackage[accsupp]{axessibility}

\newcolumntype{C}{>{\centering\arraybackslash}X}
\newcolumntype{L}{>{\raggedright\arraybackslash}X}
\newcolumntype{R}{>{\raggedleft\arraybackslash}X}

\newcommand{\mysquare}[1][black]{{\small\textcolor{#1}{\ensuremath\blacksquare}}}

\def\wacvPaperID{1802} 

\wacvapplicationstrack 

\wacvfinalcopy 

\ifwacvfinal
\usepackage[breaklinks=true,bookmarks=false]{hyperref}
\else
\usepackage[pagebackref=true,breaklinks=true,colorlinks,bookmarks=false]{hyperref}
\fi

\begin{document}

\title{Generative Range Imaging for Learning Scene Priors of 3D LiDAR Data}

\author{Kazuto Nakashima$^1$ \quad Yumi Iwashita$^{2}$ \quad Ryo Kurazume$^1$\\
	$^1$Kyushu University, Fukuoka, Japan\\
	$^2$Jet Propulsion Laboratory, California Institute of Technology, Pasadena, CA, USA\\
	{\tt\small k\_nakashima@irvs.ait.kyushu-u.ac.jp \quad yumi.iwashita@jpl.nasa.gov \quad kurazume@ait.kyushu-u.ac.jp}
}

\maketitle
\thispagestyle{empty}

\begin{abstract}
	3D LiDAR sensors are indispensable for the robust vision of autonomous mobile robots. However, deploying LiDAR-based perception algorithms often fails due to a domain gap from the training environment, such as inconsistent angular resolution and missing properties. Existing studies have tackled the issue by learning inter-domain mapping, while the transferability is constrained by the training configuration and the training is susceptible to peculiar lossy noises called ray-drop. To address the issue, this paper proposes a generative model of LiDAR range images applicable to the data-level domain transfer. Motivated by the fact that LiDAR measurement is based on point-by-point range imaging, we train an implicit image representation-based generative adversarial networks along with a differentiable ray-drop effect. We demonstrate the fidelity and diversity of our model in comparison with the point-based and image-based state-of-the-art generative models. We also showcase upsampling and restoration applications. Furthermore, we introduce a Sim2Real application for LiDAR semantic segmentation. We demonstrate that our method is effective as a realistic ray-drop simulator and outperforms state-of-the-art methods.
\end{abstract}

\section{Introduction}
\label{sec:introduction}

A LiDAR sensor is a laser-based range sensor that can measure the surrounding geometry as a 3D point cloud. 
Compared to other depth cameras and radars, LiDAR sensors cover a wide field of view and are also robust to lighting conditions owing to their active sensing based on the pulsed laser. 
Therefore, LiDAR-based 3D perception has become an indispensable component of autonomous mobile robots and vehicles.

\begin{figure}[tb]
	\centering
	\includegraphics[width=\hsize]{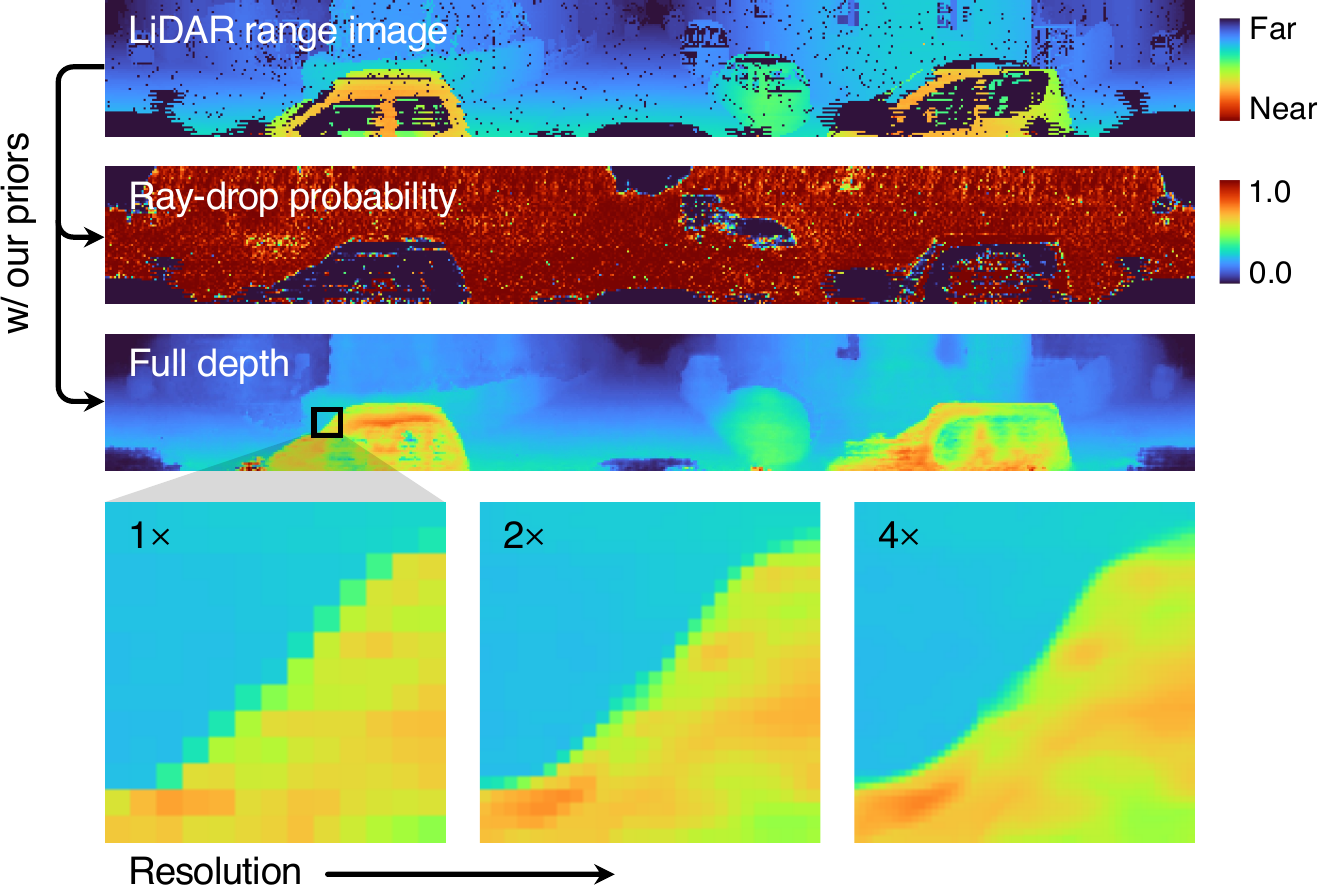}
	\caption{LiDAR range imaging (top) involves the missing points spreading unevenly, called ray-drops. Our method enables us to infer the ray-drop probability and the underlying complete scene (middle) on the continuous image representation (bottom).}
	\label{fig:introduction}
\end{figure}

In particular, semantic segmentation on LiDAR point clouds~\cite{qi2017pointnet,qi2017pointnet2,wu2018squeezeseg,wu2019squeezesegv2,xu2020squeezesegv3,zhao2021epointda,milioto2019rangenet++} is one of the most important tasks in autonomous navigation, which identifies 3D objects at the point level for traversability estimation.
For general point clouds, the most recent segmentation methods are based on PointNet~\cite{qi2017pointnet} and PointNet++~\cite{qi2017pointnet2}, neural network architectures designed to deal with the unordered nature of point clouds.
However, they have limitations regarding computational speed~\cite{xu2020squeezesegv3} and memory requirements~\cite{behley2019iccv} for large-scale point clouds and are often performed on a reduced size.
To further efficient training and inference, the spherical projection has been exploited in the LiDAR segmentation.
In this approach, point clouds are represented as a bijective 2D grid, a so-called \textit{range image} (see Fig.~\ref{fig:introduction} for instance).
So far, many studies~\cite{wu2018squeezeseg,wu2019squeezesegv2,xu2020squeezesegv3,zhao2021epointda,milioto2019rangenet++} proposed 2D convolutional neural networks that perform point cloud segmentation on this range image representation.

While the range image representation has improved the processing efficiency, there are domain gap problems that degrade the performance in deploying trained models.
In this paper, we argue two issues relating to ray-casting and ray-dropping.
The ray-casting issue is derived from the angular configuration of emitted lasers. LiDAR sensors have a wide variety of angular resolutions due to hardware constraints, which might bring spatial bias in training the segmentation models.
The ray-dropping issue is derived from the phenomenon of laser reflection. 
While LiDAR sensors are robust to day-and-night illumination changes, produced range images involve quite a few missing points if the reflected laser intensity is too low due to scene-derived specular and diffuse reflection and light absorption.
This artifact is problematic in simulation-to-real (Sim2Real) segmentation tasks because the phenomenon is non-trivial to be reproduced in a simulator.
Some studies proposed LiDAR domain adaptation methods to address the ray-casting~\cite{langer2020domain,yi2021complete} and ray-dropping issues~\cite{wu2019squeezesegv2,wu2018squeezeseg,zhao2021epointda,manivasagam2020lidarsim}.

In this work, we propose a generative model-based method for LiDAR domain adaptation.
Our method learns the generative process of LiDAR range images alongside the ray-casting and ray-dropping effects in an end-to-end manner, based on generative adversarial networks (GANs)~\cite{goodfellow2014generative}.
The learned data priors can be used for mapping different domains.
Our model builds upon the recently proposed two paradigms for generative models: implicit neural representation~\cite{Skorokhodov_2021_CVPR,anokhin2021image} and lossy measurement model~\cite{bora2018ambientgan,kaneko2020noise,nakashima2021learning}. 
Implicit neural representation is a continuous and differentiable signal representation parameterized by neural networks. 
For example, an image is represented by a coordinate-based function and its resolution is determined by coordinate queries. 
Motivated by this scheme, we aim to model the editable ray-casting process of LiDAR range images. 
The lossy measurement model is an invertible function that simulates the stochastic signal corruption along the data generation. 
We aim to model the scene-dependent ray-dropping in an unsupervised manner.

In Section~\ref{sec:fidelity_and_diversity}, we first evaluate our model in terms of generation fidelity and diversity compared to the point-based and image-based state-of-the-art generative models. 
Our model shows the best results in most standard image/point cloud metrics. We also examine the validity of feature-based metrics on LiDAR point clouds, motivated by de facto standard assessment in the natural image domain~\cite{NIPS2017_8a1d6947}. 
We then showcase applications with our model such as post-hoc upsampling and data restoration from sparse depth observation. 
Finally, in Section~\ref{sec:sim2real}, we conduct Sim2Real semantic segmentation by using our model as a noise simulator. 
We demonstrate that our method produces realistic ray-drop noises and outperforms the state-of-the-art LiDAR Sim2Real methods.
In summary, our contributions are as follows:
\begin{itemize}
	\item We propose a novel GAN for LiDAR range images simulating the ray-casting and ray-dropping processes. 
	\item We showcase the utility of our model on the post-hoc upsampling and data restoration.
	\item We apply our model to Sim2Real semantic segmentation. We empirically show that our model produces realistic ray-drop noises on simulation data and outperforms the state-of-the-art methods.
\end{itemize}

\section{Related work}
\label{sec:related_work}

\subsection{LiDAR domain adaptation}

As in the field of natural images, the performance of LiDAR perception tasks also suffers from domain shift problems between the training and test environments~\cite{triess2021survey} such as by different sensor configurations, geography, weather conditions, and simulation.
We highlight the following two cases focused on in this work.

\noindent
\textbf{Angular resolution.}
Sampling angular resolution is a non-negligible property of LiDAR sensors, which determine the density of 3D point clouds.
To mitigate the gap, Langer \textit{et al.}~\cite{langer2020domain} used pseudo LiDAR range images sampled from sequentially superimposed point clouds or meshes.
However, the synthesis quality may depend on the sequential density of scans. 
Yi \textit{et al.}~\cite{yi2021complete} proposed voxel-based completion to bridge the scan-wise discrepancy, while this approach is designed for point-based perception methods.
This paper proposes a scan-wise transfer using GAN-based data prior and shows some qualitative results. 

\noindent
\textbf{Ray-drop noises.}
Ray-dropping occurs if the pulsed lasers failed to reflect from measured objects.
This phenomenon is caused by complex physical factors such as mirror diffusion, specular diffusion, light absorption, and range limits.
In the aspect of perception tasks, the ray-drop noises are one of the important properties of \textit{real} data~\cite{wu2018squeezeseg,wu2019squeezesegv2,zhao2021epointda,manivasagam2020lidarsim}.
Some studies tackled simulating the ray-drop noises to make the LiDAR simulator realistic.
SqueezeSeg~\cite{wu2018squeezeseg} proposed to sample binary noises based on the pixel-wise frequency computed from the real data.
However, the averaged noises cannot make object-wise effects so they might be far from the actual distribution.
To estimate ray-drop noises from LiDAR range images, Zhao \textit{et al.}~\cite{zhao2021epointda} trained CycleGAN~\cite{zhu2017unpaired} and Manivasagam \textit{et al.}~\cite{manivasagam2020lidarsim} trained U-Net~\cite{ronneberger2015u}.
However, those approaches cast the task as a binary classification based on a cross-entropy objective.
As mentioned by Manivasagam \textit{et al.}~\cite{manivasagam2020lidarsim}, the cross-entropy training does not guarantee the estimated probability is calibrated.
We hypothesize that such approximated noise simulation might be suboptimal performance in Sim2Real tasks.

\subsection{Deep generative modeling}

In recent years, great progress has been achieved in generative models based on deep neural networks.
In particular, a generative adversarial network (GAN)~\cite{goodfellow2014generative} has been attracting a great deal of attention in the image domain due to their sampling quality and efficiency~\cite{bond2021deep}.
As an example of recent studies, Karras \textit{et al.}~\cite{karras2018progressive} proposed ProGAN for synthesizing mega-pixel natural images, and the generation quality has been significantly improved in a few years~\cite{karras2019style,karras2020analyzing,karras2020training}.
Moreover, well-trained GANs can be used as \textit{generative image priors} for semantic manipulation and data restoration~\cite{gu2020image,roich2021pivotal}.

\noindent
\textbf{Implicit neural representation.}
A GAN has also taken another step forward with implicit neural representation~\cite{anokhin2021image,Skorokhodov_2021_CVPR}.
Implicit neural representation is a method of continuous signal representation using a coordinate-based neural network.
The typical architecture employs MLPs that receive arbitrary coordinate points and predict values such as signed distances~\cite{park2019deepsdf} for 3D shapes, colors~\cite{anokhin2021image,Skorokhodov_2021_CVPR} for 2D images, and colors/density~\cite{schwarz2020graf} for volumetric rendering.
This implicit scheme allows the models to learn the resolution-independent representation even if trained with discretized data such as images.
CIPS~\cite{anokhin2021image} and INR-GAN~\cite{Skorokhodov_2021_CVPR} incorporated the idea into image GANs.
They demonstrated that the models could control the resolution to perform spatial interpolation and extrapolation.
This paper discusses the availability in modeling ray-casting.

\noindent
\textbf{Lossy measurement models.}
Training datasets are not always \textit{clean} and can be problematic on training GANs.
Since the objective of GANs is to mimic a distribution of a dataset, the learned generator space can also be noisy.
To address the issue, some studies~\cite{bora2018ambientgan,kaneko2020noise,nakashima2021learning} introduced a \textit{probabilistically} invertible function into the generative process to learn clean signals from only the noisy datasets.
For instance on multiplicative binary noises, Bora \textit{et al.}~\cite{bora2018ambientgan} proposed AmbientGAN and Li \textit{et al.}~\cite{li2018learning} proposed MisGAN.
Their noise models are formulated with a signal-independent probability, which does not meet the LiDAR's signal-dependent binary noises.
Kaneko and Harada~\cite{kaneko2020noise} proposed NR-GAN which can estimate the signal-dependent noise distribution; however binary noises are not covered.

\noindent
\textbf{LiDAR applications.}
Although most generative models have focused on the natural image datasets, several studies~\cite{caccia2019deep,nakashima2021learning} have begun to apply them to LiDAR data.
Caccia \textit{et al.}~\cite{caccia2019deep} proposed the first application of deep generative models on LiDAR data. 
They trained the variational autoencoders (VAEs)~\cite{kingma2013auto} on the range image representation.
They also reported the visual results by a vanilla GAN~\cite{radford2015unsupervised}. 
However, the distribution of LiDAR range images can be discrete due to the random ray-drop noises, which is difficult to be represented by neural networks as a continuous function.
Motivated by the concept of the lossy measurement models, Nakashima and Kurazume~\cite{nakashima2021learning} proposed DUSty incorporating the differentiable ray-drop effect into GANs to robustly train the LiDAR range images.
They employed straight-through Gumbel-Sigmoid distribution~\cite{maddison2017concrete,Jang2017categorical} for modeling the LiDAR noises, so that the model learns the discrete data distribution as a composite of two modalities: an underlying complete depth and the corresponding uncertainty of measurement.
Those two studies empirically showed the availability of generative models on LiDAR range images, while did not evaluate the effectiveness on the actual perception tasks.
In contrast, our paper improves the generation quality and also provides the quantitative evaluation on Sim2Real semantic segmentation.

\section{Method}
\label{sec:method}

\begin{figure}[tb]
	\footnotesize
	\centering
	\includegraphics[width=0.95\hsize]{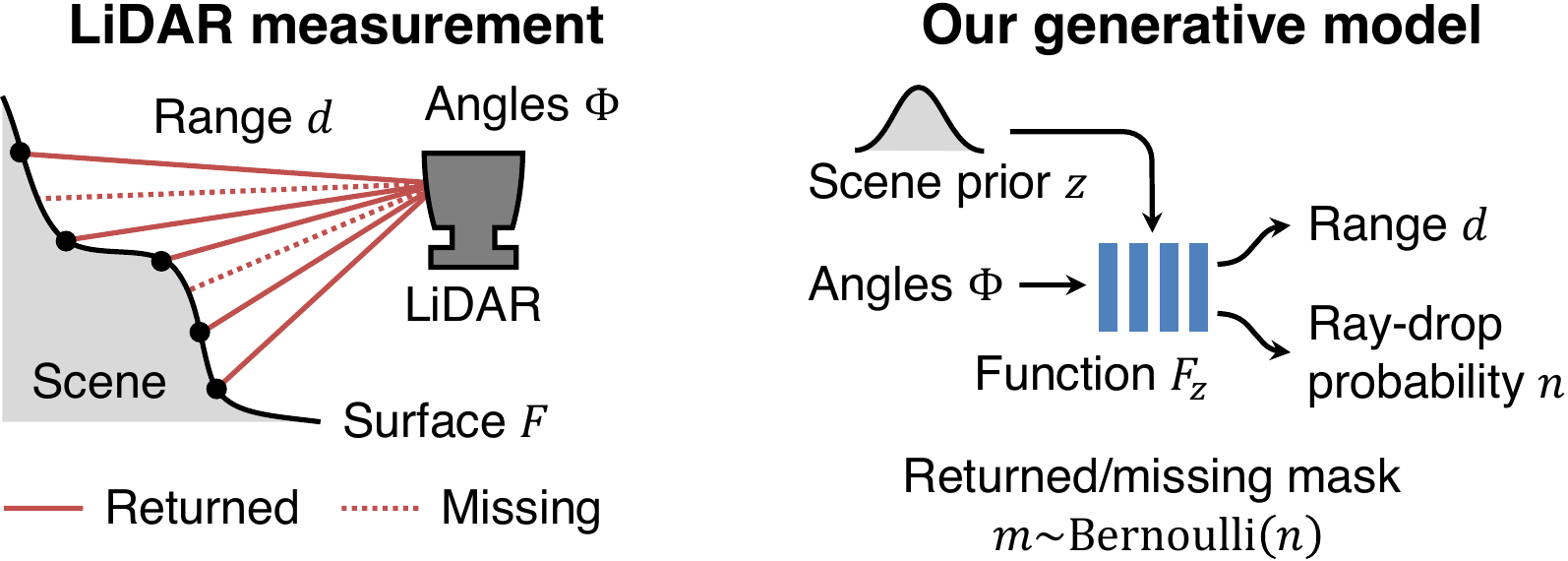}
	\\ \vspace{3mm}
	\caption{A schematic diagram of LiDAR measurement and our proposed generative model for 3D LiDAR data. We hypothesize that laser dropping occurs stochastically according to scene-specific probabilities.}
	\label{fig:introduction:summary}
\end{figure}

Fig.~\ref{fig:introduction:summary} depicts the LiDAR measurement and our formulation with deep generative models.
Our aim is to learn the LiDAR scene priors independent of angular resolution and to leverage them for data-level domain transfer.
To this end, Section~\ref{sec:method:implicit_representation} first introduces a resolution-free implicit representation of range images. 
In Section~\ref{sec:method:generative_model}, we then present our GAN based on the implicit representation and differentiable ray-dropping effect. 
Finally, Section~\ref{sec:method:inference} introduces an inference step which uses our learned GAN as generative scene priors.
We provide the implementation details in the supplementary materials.
Our code is available at \url{https://github.com/kazuto1011/dusty-gan-v2}.

\subsection{Implicit representation of range images}
\label{sec:method:implicit_representation}

\noindent
\textbf{LiDAR range images.}
While a general representation of point clouds is a set of Cartesian coordinate points $\{(p_x,p_y,p_z)\}$~\cite{achlioptas2018learning,qi2017pointnet,yang2019pointflow}, LiDAR point clouds can also be represented as a bijective 2D grid~\cite{nakashima2021learning,caccia2019deep,triess2020iv,zhao2021epointda,wu2018squeezeseg,wu2019squeezesegv2} due to the measurement mechanism, \textit{range imaging}.
Suppose that a LiDAR sensor emitting horizontal $W$ pulsed lasers for $H$ elevation angles measures a distance $d$ for each angular position. 
Then all the distance values can be assigned to a $H\times W$ angular grid by spherical projection, where the resultant representation is called a range image.
Each pixel has a set of sensor-dependent azimuth and elevation angles $\Phi=(\theta, \phi)$ and the corresponding distance $d$.
Therefore, an arbitrary LiDAR scene $\{(p_x,p_y,p_z)\}$ can be seen as a set of spherical coordinates $\{(\theta, \phi, d)\}$ projected on the 2D grid.

\noindent
\textbf{Scenes as a function.}
The core idea of this paper is to represent the 3D scene by a function $F$ mapping the spherical angular set to the distance: $d=F(\Phi)$.
If the scene can be represented in the continuous function space $F$, we can reconstruct the scene with arbitrary resolution queries.
Furthermore, it is possible to express multiple scenes by introducing parameters $z$ that condition the function $F$.
To this end, we aim to build the function $d=F(\Phi, z)$ as a deep generative model where the scene is instantiated by the sensor-agnostic scene prior $z$ and queried by the sensor-specific angular set $\Phi$ to generate the LiDAR range images.

\begin{figure*}[tb]
	\centering
	\includegraphics[width=0.9\hsize]{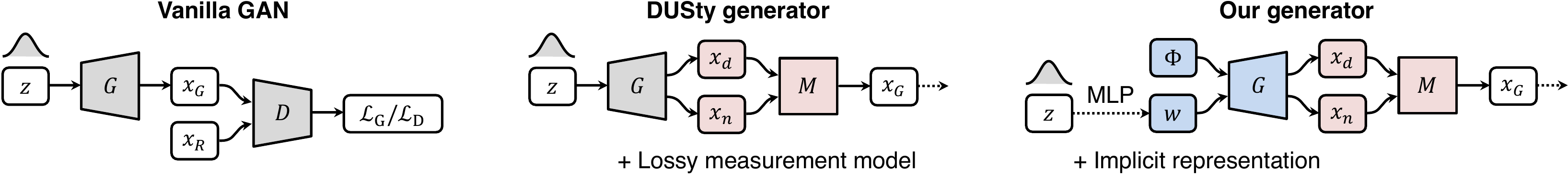}
	\caption{Comparison of the baseline GANs and ours. The vanilla GAN~\cite{caccia2019deep} consists of the generator $G$ and the discriminator $D$ and directly learns the distribution of raw range images $x_R$. DUSty generator~\cite{nakashima2021learning} disentangles the generator space as the range $x_d$ and measurability $x_n$ (ray-drop probability) with the self-conditioned measurement model $M$. Our generator further introduces implicit neural representation~\cite{Skorokhodov_2021_CVPR} so that the spatial resolution is not fixed by the generator but controlled by the external query $\Phi$.}
	\label{fig:model}
\end{figure*}

\subsection{Generative range imaging}
\label{sec:method:generative_model}

\noindent
\textbf{Generative adversarial networks.}
To introduce the latent variable $z$ that controls the scene, we build the function $F$ as a generative model.
We employ a generative adversarial network (GAN)~\cite{goodfellow2014generative}, similar to prior work~\cite{nakashima2021learning,caccia2019deep}.
A GAN typically consists of two networks: a generator $G$ and a discriminator $D$.
In image synthesis tasks, $G$ maps a latent variable ${z}\sim N(0,{I})$ to an image $x_G=G({z})$, whereas $D$ tells the generated image $x_G$ from sampled real images $x_R$.
The networks are trained in an alternating fashion by minimizing the adversarial objective, \textit{e.g.}, the following non-saturating loss~\cite{goodfellow2014generative}:
\begin{eqnarray}
	\label{eq:adversarial_d}
	\mathcal{L}_D=-\mathbb{E}_x[\log D({x}_R)]-\mathbb{E}_z[\log (1-D(G({z})))], \\
	\label{eq:adversarial_g}
	\mathcal{L}_G=-\mathbb{E}_z[\log D(G({z}))].
\end{eqnarray}

In this paper, the generator $G$ is equivalent to the aforementioned function $F$ and the structure of $x_G$ is represented by scene condition $z$ and given coordinates $\Phi$.
Our GAN builds upon INR-GAN~\cite{Skorokhodov_2021_CVPR}, which was demonstrated on natural images.
INR-GAN first transforms the latent variable $z$ to disentangled style space $w$ by MLPs and modulates the network weights to synthesize images.

\noindent
\textbf{Lossy measurement model.}
\label{sec:lossy}
LiDAR range images involve many missing points caused by ray-dropping.
In the aspect of training GANs, the missing points prevent stability and fidelity of depth surfaces.
To address this issue, we combine our model with the ray-dropping formulation proposed in DUSty~\cite{nakashima2021learning}.
They assumes the ray-dropping phenomenon is stochastic and uses Bernoulli sampling with self-conditioned probability.
As outputs from the generator $G$, DUSty first assumes a complete range image $x_d\in\mathbb{R}^{H\times W}$ and the corresponding ray-drop probability map $x_n\in\mathbb{R}^{H\times W}$.
Then, the final LiDAR measurement $x_G$ is sampled from the complete $x_d$ according to a lossy measurement mask $m \sim \mathrm{Bernoulli}\left(x_n\right)$.
Since the sampling $m$ is non-differentiable, $m$ is reparameterized with the straight-through Gumbel-Sigmoid distribution~\cite{Jang2017categorical} to estimate the gradients.
Note that the generator spaces $x_d$ and $x_n$ do not have to be discrete distribution, while $x_G$ can produce discrete noises caused by ray-dropping.
As in DUSty, we convert each distance value $x_d$ into the \textit{inverse} depth for further stable training.
In Fig.~\ref{fig:model}, we compare the vanilla GAN~\cite{caccia2019deep}, DUSty~\cite{nakashima2021learning}, and our proposed model.

\noindent
\textbf{Positional encoding for circular grid.}
In the fields of implicit neural representation~\cite{Skorokhodov_2021_CVPR,anokhin2021image}, positional encoding is an indispensable technique to represent high-frequency details of output images, which transforms coordinates points to high-dimensional feature space.
In particular, Fourier features~\cite{tancik2020fourier} are the most popular encoding scheme in the image domain where the coordinate $\Phi=(\theta, \phi)$ is transformed by the following sinusoidal function:
\begin{eqnarray}
	\mathrm{PE}(\theta,\phi) = \sin([b_\theta,b_\phi][\theta,\phi]^\top),
\end{eqnarray}
where $b_\theta\in\mathbb{R}^D$ and $b_\phi\in\mathbb{R}^D$ are weight vectors which control frequencies in encoded space, and $\theta, \phi \in [-\pi,\pi]$ are angular values determined by LiDARs.
In natural image applications, the weights can be set by various formula such as a power of two~\cite{mildenhall2020nerf}, Gaussian samples~\cite{tancik2020fourier}, and learnable parameters~\cite{anokhin2021image,Skorokhodov_2021_CVPR}.
In our case, the weights should be carefully initialized so that the encoding preserves the cylindrical structure of the angular input. 
First, we set the limit value of output frequencies both for azimuth and elevation inputs. We then uniformly sample $b_\phi$ within the determined limit and sample $b_\theta$ from a set of power of two.

\noindent
\textbf{Subgrid training.}
The sampled frequencies in positional encoding are sparse in the horizontal direction.
If the subsequent blocks are overfitted with a small number of fixed sampling points, post-hoc changes or upsampling in the angular input may result in unexpected aliasing at intermediate layers.
This is not compatible with our purpose to control the generation format according to the sensor specification.
To this end, we augment the angular inputs by a random phase shift horizontally during training and translate the output images in inverse direction for the number of corresponding pixels.

\subsection{Domain-agnostic inference}
\label{sec:method:inference}

\begin{figure}[t]
	\centering
	\includegraphics[width=\hsize]{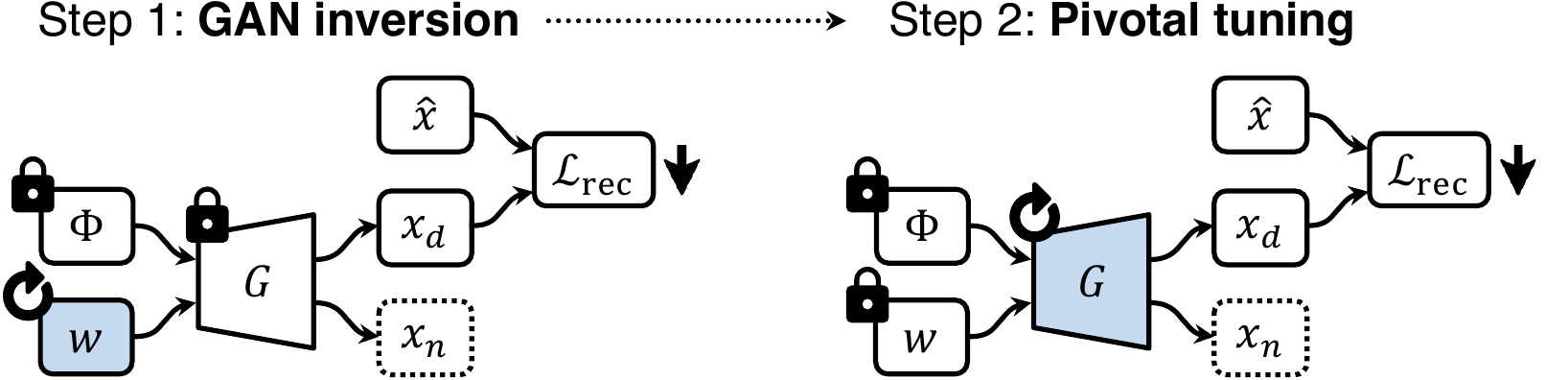}
	\vspace{1mm}
	\caption{Overview of our inference method based on PTI~\cite{roich2021pivotal}. Step 1 optimizes the latent code $w$ and step 2 fine-tunes weights $\Omega$ of the generator $G$. The by-product $x_n$ can be used for Sim2Real applications.}
	\label{fig:inversion}
\end{figure}

This section describes a method to reconstruct and compensate for the arbitrary LiDAR measurement using \textit{GAN inversion}~\cite{xia2022gan}.
The GAN inversion is an inference task to find the latent representation of the given data.
The standard approaches for GAN inversion have two camps: auto-decoding~\cite{roich2021pivotal, karras2020analyzing} which optimizes the latent codes to match the data, and training additional encoders~\cite{Perarnau2016} to estimate the latent code directly.
In this paper, we employ Pivotal Tuning Inversion (PTI)~\cite{roich2021pivotal}, one of the latest auto-decoding approaches.
This section briefly introduce its steps along with our designed optimization objective.
Fig.~\ref{fig:inversion} shows the overview of our method.

\noindent
\textbf{Step 1: GAN inversion.}
As the latent representation, we use the aforementioned style code $w$ instead of $z$.
Let $\hat{x}$ and $\hat{m}$ are a target depth map and the corresponding ray-drop mask.
We first define the following objective to assess the depth error:
\begin{eqnarray}
	\mathcal{L}_{\rm{rec}}=\frac{ \lVert \hat{m} \odot (1-x_d(w,\Phi;\Omega)/\hat{x}) \rVert_{1} }{\lVert \hat{m} \rVert_{1}},
\end{eqnarray}
where $x_d(w,\Phi;\Omega)$ is a generated depth map conditioned by the latent code $w$, resolution query $\Phi$, and generator weights $\Omega$. 
The objective $\mathcal{L}_{\rm{rec}}$ measures the relative absolute error normalized by the valid points in $\hat{m}$.
In this step, we compute the matching code by $\hat{w}=\arg \min_{w} \mathcal{L}_{\rm{rec}}$.

\noindent
\textbf{Step 2: pivotal tuning.}
With the optimized $\hat{w}$, we can generate a range image $x_G$ similar to the target $\hat{x}$.
This step further minimize the minor appearance difference by fine-tuning the generator weights $\Omega$ while freezing the pre-optimized code $\hat{w}$ to pivot on the fundamental structure.
We minimize the same objective with respect to $\Omega$: $\hat{\Omega}=\arg \min_{\Omega} \mathcal{L}_{\rm{rec}}$.

Since our masked objective $\mathcal{L}_{\rm{rec}}$ relies on only measurable points, we can also apply the inference to restore partially observed images.
Moreover, we can obtain the by-product $x_n$ by reconstructing simulation data as $\hat{x}$, which can be used to simulate the ray-drop noises.

\section{Evaluation}

\subsection{Generation fidelity and diversity}
\label{sec:fidelity_and_diversity}

Following the related studies on generative models, we first evaluate fidelity and diversity of generated samples.

\noindent
\textbf{Dataset.}
We use the KITTI Raw dataset~\cite{geiger2013vision} since the other popular releases such as KITTI Odometry~\cite{geiger2013vision} include missing artifacts by ego-motion correction~\cite{triess2020iv}.
The dataset provides 22 trajectories of scans measured by the Velodyne HDL-64E LiDAR, where each scan has 64 layers vertically.
For future derivative work, we use standard splits defined by KITTI Odometry~\cite{geiger2013vision}\footnote{We use 18,329 scans for training (sequence 3 is not available), 4,071 scans for validation from ``city'', ``road'', and ``residential'' categories. We define a test set by the remaining 18,755 scans since the correspondence is not publicly available.}.
The provided data are sequences of Cartesian coordinate points ordered by angles.
We first chunk the ordered sequence into 64 sub-sequences, where each represents one elevation angle.
We then subsample 512 points for each sub-sequence and stack them to form a $64\times512$ range image.

\noindent
\textbf{Baselines.}
We compare our model with two popular baselines of point-based generative models: r-GAN~\cite{achlioptas2018learning} and l-WGAN~\cite{achlioptas2018learning}.
r-GAN is a kind of GANs based on point set representation, which consists of an MLP generator and a PointNet~\cite{qi2017pointnet} discriminator.
l-WGAN first learns a PointNet-MLP autoencoder and then performs adversarial training with the additional MLPs to generate the bottleneck feature.
We train l-WGAN autoencoder by measuring the earth mover's distance (EMD) between input and reconstruted point clouds.
The size of LiDAR point clouds is much larger than the typical benchmark for point-based methods and the EMD distance calculation takes extremely high computational cost both in training and evaluation.
For efficiency and fareness, we first downsample the LiDAR point clouds to the conventional number of points 2048 by farthest point sampling (FPS).
We also compare with two types of image-based generative models: a vanilla GAN~\cite{caccia2019deep} and DUSty~\cite{nakashima2021learning}.
The vanilla GAN and DUSty share the backbone design based on $4\times4$ transposed convolution, while DUSty is adopted the ray-drop measurement model explained in Section~\ref{sec:lossy}. 
Those models cannot change the output resolution from the training configuration.

\noindent
\textbf{Point-based metrics.}
Following the related work on the point-based generative models~\cite{achlioptas2018learning,yang2019pointflow,nakashima2021learning}, we measured four types of distributional similarities between the sets of reference and generated point clouds, defined by Yang~\textit{et al.}~\cite{yang2019pointflow}: Jensen--Shannon divergence (JSD) for fidelity, coverage (COV) for diversity, minimum matching distance (MMD) for fidelity, and 1-nearest neighbor accuracy (1-NNA) for both fidelity and diversity evaluation.
To compute the distance between point clouds for COV, MMD, and 1-NNA, we used EMD.

\noindent
\textbf{Image-based metrics.}
Additionally, we computed the sliced Wasserstein distance (SWD)~\cite{karras2018progressive} to measure the patch-based image similarity for evaluating quality of the inverse depth maps.
SWD is computed based on $7\times7$ patches extracted from three level of image pyramids.
For all metrics, we report the mean scores over three runs with different seeds.

\begin{figure}[t]
	\includegraphics[width=0.95\hsize]{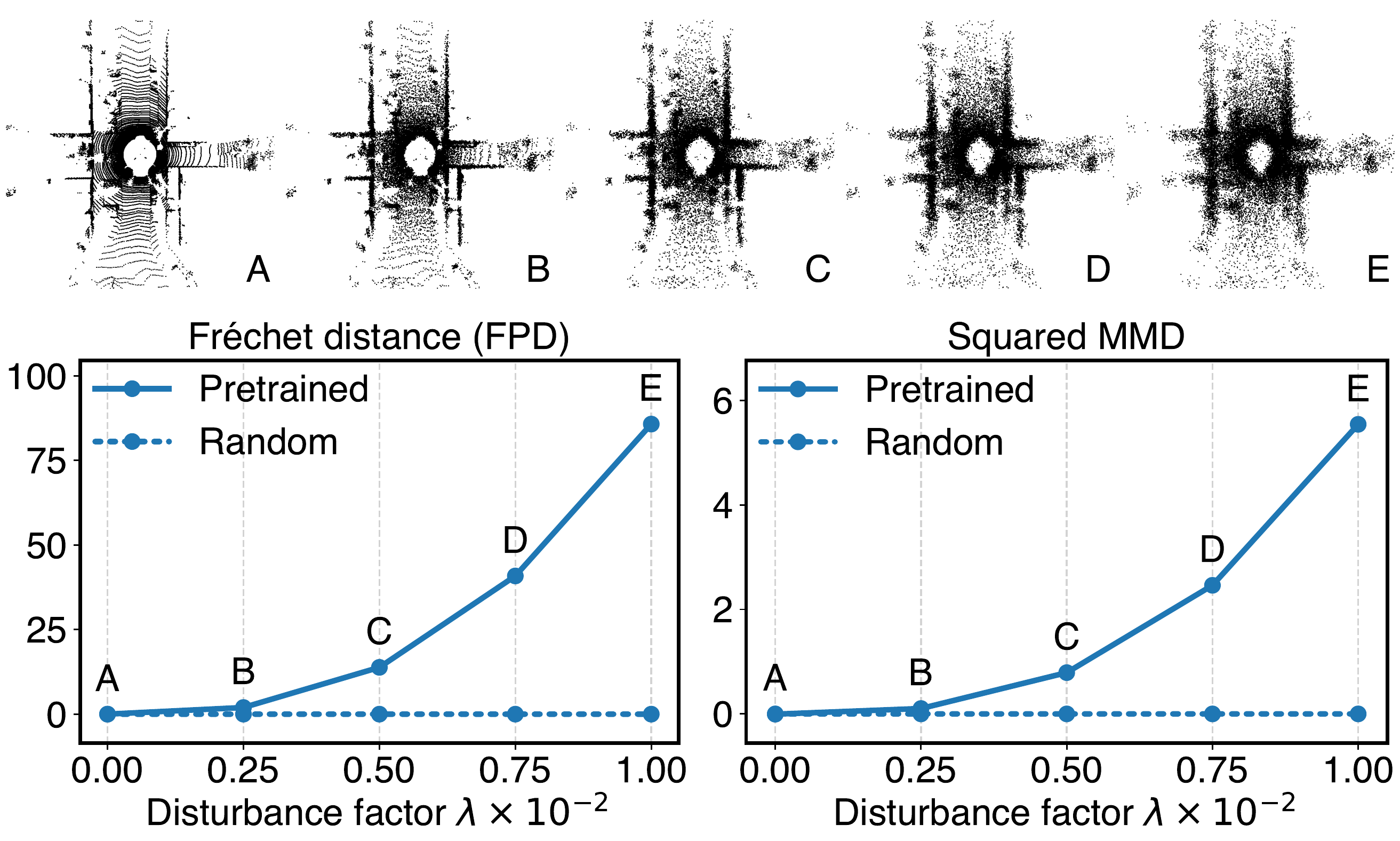}
	\caption{Sanity checks of the Fr\'{e}chet distance (FPD~\cite{Shu_2019_ICCV}) and the squared MMD~\cite{binkowski2018demystifying} on feature representation of LiDAR point clouds. We applied additive Gaussian noises with a coefficient $\lambda$ to the KITTI point clouds (see A--E) and computed the metrics with the \textit{clean} original point clouds. All point clouds were encoded by PointNet~\cite{qi2017pointnet} with pre-trained or random weights.}
	\label{fig:sanity_check}
\end{figure}

\noindent
\textbf{Feature-based metrics.}
Existing evaluation metrics~\cite{yang2019pointflow} for synthesized point clouds is impracticable for large number of points due to the limited scalability of sample-to-sample distance computation such as EMD.
Nakashima \textit{et al.}~\cite{nakashima2021learning} addressed the issue on LiDAR point clouds by subsampling real and generated point clouds in evaluation. 
As an alternative metrics, Shu \textit{et al.}~\cite{Shu_2019_ICCV} proposed Fr\'{e}chet point cloud distance (FPD) as an analogy of FID~\cite{NIPS2017_8a1d6947} which is the de facto standard metrics in the image domain.
FPD first maps an arbitrary number of points to a low-dimensional feature space by PointNet pre-trained on ShapeNet~\cite{shapenet2015}, then measures the Fr\'{e}chet distance between the real and generated data distributions in the feature space. 
Unlike the aforementioned point cloud metrics, even large-scale point clouds such as LiDAR data can be calculated without downsampling. 
Following that Heusel \textit{et al.}~\cite{NIPS2017_8a1d6947} evaluated the sensitivity of FID under various image disturbances, we first confirm the validity of the off-the-shelf PointNet on LiDAR point clouds. 
We apply additive Gaussian noises with a coefficient $\lambda$ to the KITTI training set and compute FPD against the clean original training set.
On PointNet features, we also compute squared maximum mean discrepancy (squared MMD)~\cite{binkowski2018demystifying} which is also used in the image domain as KID.
Fig.~\ref{fig:sanity_check} shows that as the strength of the perturbation increases, both metrics reflect the dissimilarity.
The further results under various disturbance effects are provided in our supplementary material.

\noindent
\textbf{Quantitative comparison.}
Table~\ref{tab:evaluation_pointnet} reports FPD and squared MMD.
We can see that our method outperformed both the point-based and image-based baselines with large margins.
Table~\ref{tab:evaluation_swd} reports SWD for image-based methods.
Our method showed the best performance even in this 2D level assessment.
Finally, Table~\ref{tab:evaluation_pointcloud} reports JSD, COV, MMD, and 1-NNA.
Except for JSD and MMD, the proposed method showed the highest score.
As for l-WGAN showing the best in MMD, we believe this is due to the fact that it optimizes EMD directly in training of the autoencoder.
We provide a generated samples of all methods in the supplementary materials.

\begin{table}[t]
	\centering
	\footnotesize
	\begin{threeparttable}
		\caption{Quantitative comparison by distributional similarity of PointNet features: FPD and MMD$^2$.}
		\label{tab:evaluation_pointnet}
		\begin{tabularx}{\hsize}{lRrRr}
			\toprule
			& \multicolumn{2}{c}{2048 points} & \multicolumn{2}{c}{$64\times512$ (full)} \\
			\cmidrule(lr){2-3} \cmidrule(lr){4-5}
			Method                               & FPD $\downarrow$ & MMD$^2$ $\downarrow$ & FPD $\downarrow$ & MMD$^2$ $\downarrow$ \\
			\midrule
			r-GAN~\cite{achlioptas2018learning}  & 787.45           & 45.02                & --               & --                   \\
			l-WGAN~\cite{achlioptas2018learning} & 129.35           & 10.65                & --               & --                   \\
			\midrule
			Vanilla GAN~\cite{caccia2019deep}    & 3629.36          & 671.14               & 3648.68          & 675.24               \\
			DUSty~\cite{nakashima2021learning}   & 232.90           & 39.62                & 241.32           & 42.66                \\
			\textbf{Ours}                        & \textbf{96.11}   & \textbf{3.66}        & \textbf{93.85}   & \textbf{3.84}        \\
			\bottomrule
		\end{tabularx}
	\end{threeparttable}
\end{table}

\begin{table}[t]
	\centering
	\footnotesize
	\begin{threeparttable}
		\caption{Quantitative comparison by distributional similarity of inverse depth maps: sliced Wasserstein distance (SWD $\downarrow$).}
		\label{tab:evaluation_swd}
		\begin{tabularx}{\hsize}{lCCCC}
			\toprule
			Method                             & $16\times128$  & $32\times256$  & $64\times512$  & Mean           \\
			\midrule
			Vanilla GAN~\cite{caccia2019deep}  & 0.397          & 0.371          & 0.746          & 0.505          \\
			DUSty~\cite{nakashima2021learning} & \textbf{0.353} & 0.353          & 0.768          & 0.491          \\
			\textbf{Ours}                      & 0.378          & \textbf{0.278} & \textbf{0.611} & \textbf{0.422} \\
			\midrule
			Training set                       & 0.257          & 0.207          & 0.765          & 0.410          \\
			\bottomrule
		\end{tabularx}
	\end{threeparttable}
\end{table}

\begin{table}[t]
	\centering
	\footnotesize
	\begin{threeparttable}
		\caption{Quantitative comparison by distributional similarity of point clouds: JSD$\times10^2$, COV, MMD$\times10^2$, and 1-NNA.}
		\label{tab:evaluation_pointcloud}
		\begin{tabularx}{\hsize}{lCCCC}
			\toprule
			Method                               & JSD $\downarrow$ & COV $\uparrow$ & MMD $\downarrow$ & 1-NNA $\downarrow$ \\
			\midrule
			r-GAN~\cite{achlioptas2018learning}  & 21.73            & 0.013          & 17.51            & 1.000              \\
			l-WGAN~\cite{achlioptas2018learning} & 4.91             & 0.324          & \textbf{8.62}    & 0.896              \\
			\midrule
			Vanilla GAN~\cite{caccia2019deep}    & 10.31            & 0.290          & 12.34            & 0.986              \\
			DUSty~\cite{nakashima2021learning}   & \textbf{3.00}    & 0.375          & 9.41             & 0.898              \\
			\textbf{Ours}                        & 3.04             & \textbf{0.388} & 9.12             & \textbf{0.892}     \\
			\midrule
			Training set                         & 2.80             & 0.362          & 0.765            & 0.890              \\
			\bottomrule
		\end{tabularx}
	\end{threeparttable}
\end{table}

\begin{figure}[tb]
	\centering
	\footnotesize
	\begin{tabularx}{\hsize}{CCCC}
		Target $\hat{x}$ & $1\times$ & $2\times$ & $4\times$ 
	\end{tabularx}
	\includegraphics[width=\hsize]{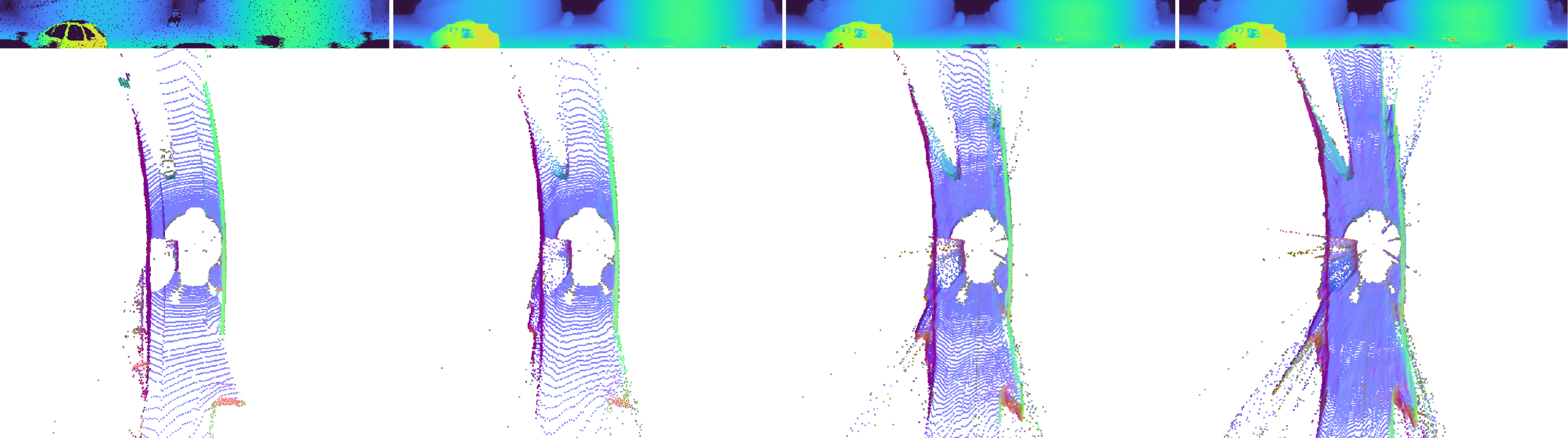}
	\caption{LiDAR data upsampling. From left to right, the target range image $\hat{x}$ and our reconstruction results $x_d$ in $1\times$, $2\times$, and $4\times$ resolutions. The bottom row shows bird's-eye views of the corresponding point clouds.}
	\label{fig:upsampling_demo}
\end{figure}

\begin{figure}[tb]
	\centering
	\footnotesize
	\setlength\tabcolsep{2pt}
	\begin{tabularx}{\hsize}{CCC}
		Original & 8 out of 64 lines & $10\%$ points \\
		\includegraphics[width=\hsize]{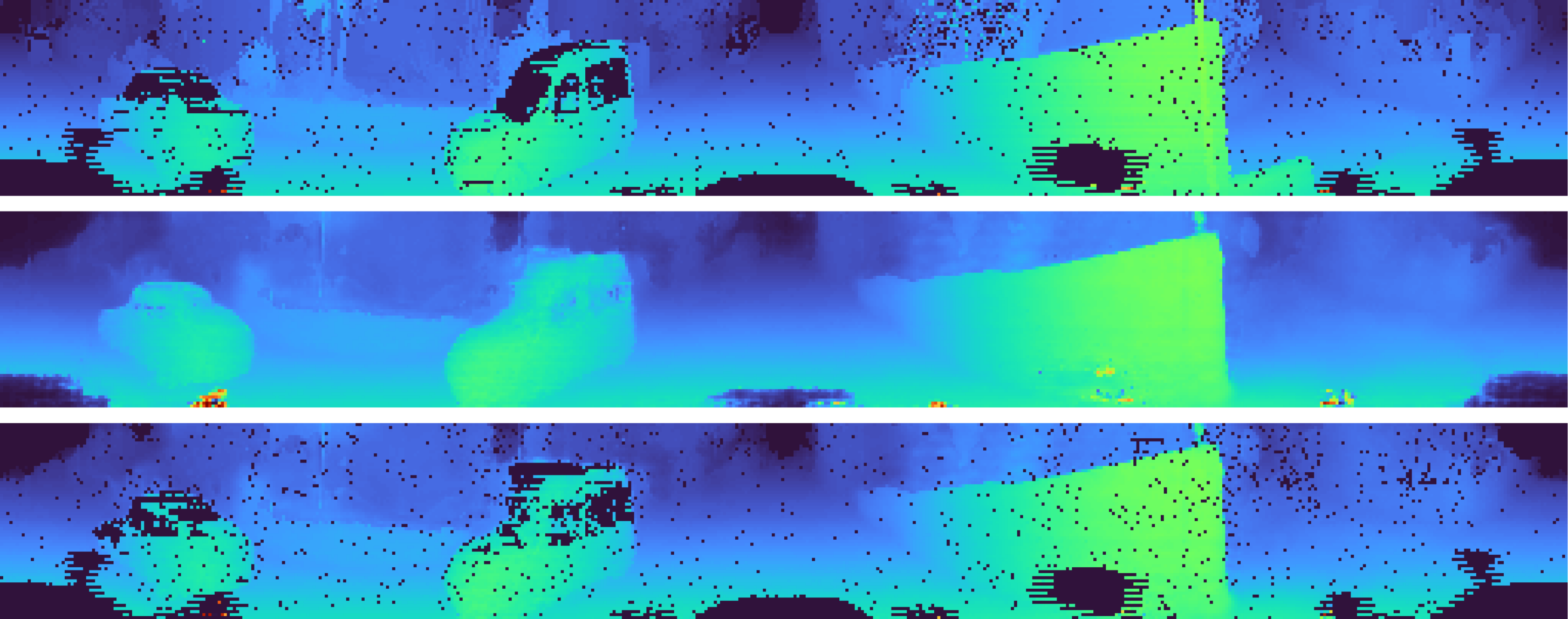} &
		\includegraphics[width=\hsize]{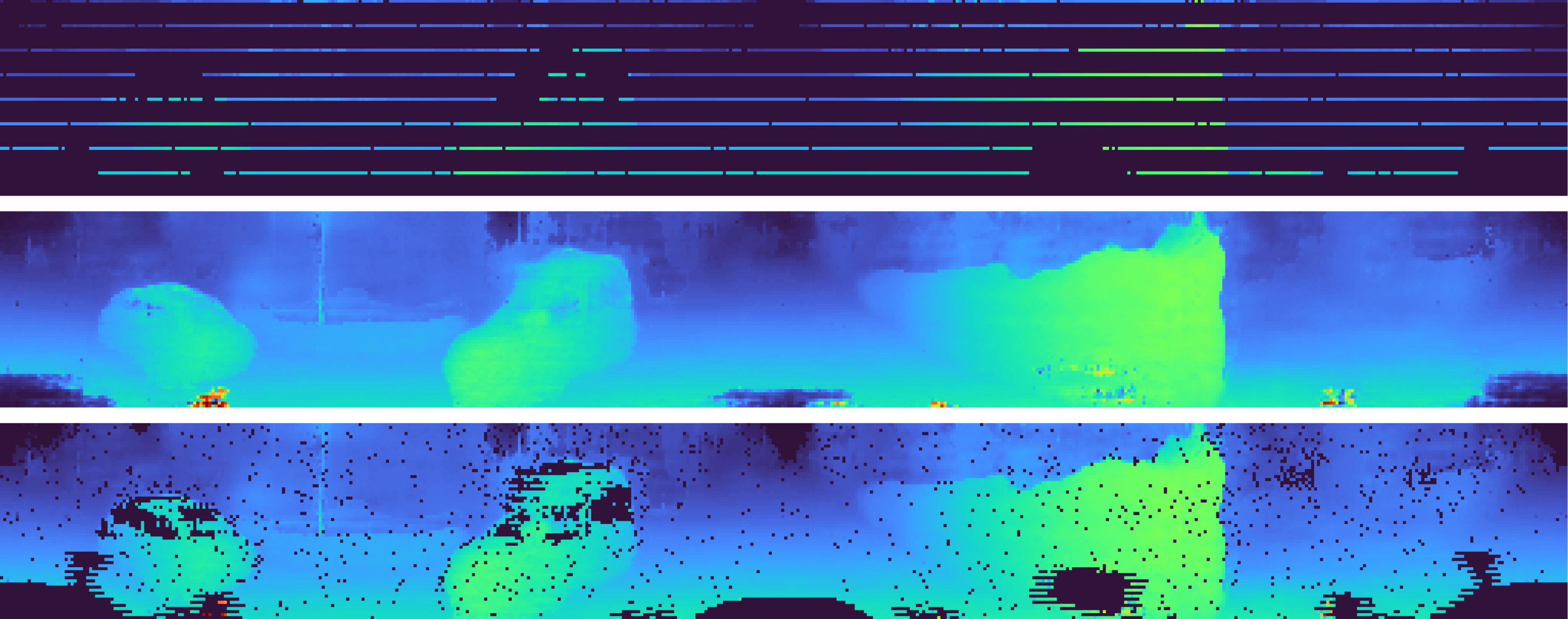} &
		\includegraphics[width=\hsize]{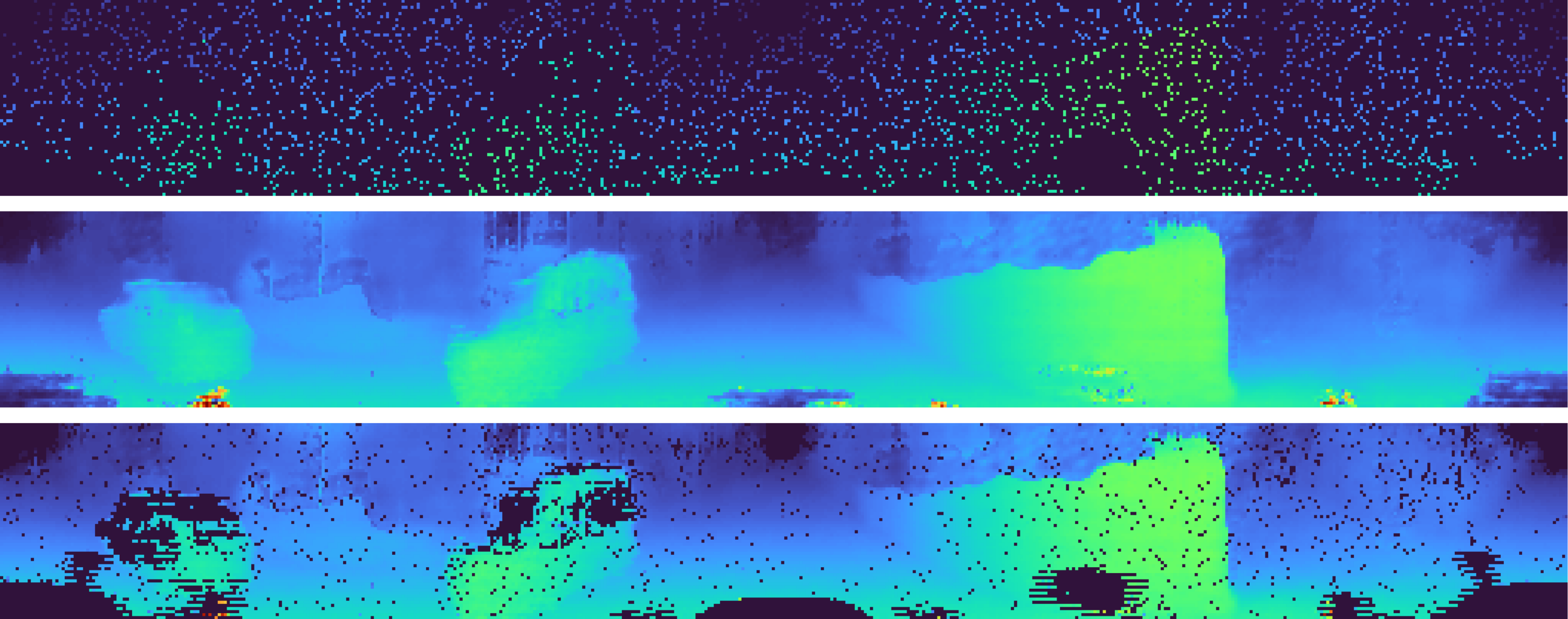}
	\end{tabularx}
	\caption{LiDAR data restoration. From top to bottom, the target range images $\hat{x}$, our reconstruction results $x_d$, and the ones with rendered ray-drops $x_G$.}
	\label{fig:restoration_demo}
\end{figure}

\noindent
\textbf{Applications.}
Figs.~\ref{fig:upsampling_demo}~and~\ref{fig:restoration_demo} show upsampling and restoration results, respectively.
Both are obtained by our auto-decoding method introduced in Section~\ref{sec:method:inference}. 
From the reasonable results, we consider our model successfully learned the scene priors of LiDAR range images.

\begin{table*}[tb]
	\centering
	\footnotesize
	\begin{threeparttable}
		\caption{Quantitative comparison of Sim2Real semantic segmentation results. We train SqueezeSegV2~\cite{wu2019squeezesegv2} on GTA-LIDAR~\cite{wu2019squeezesegv2} (simulation domain) and then evaluate precision (\%, $\uparrow$), recall (\%, $\uparrow$), and IoU (\%, $\uparrow$) on KITTI-frontal~\cite{wu2019squeezesegv2} (real domain).}
		\label{tab:sim2real_noise}
		\begin{tabularx}{\hsize}{lllCCCCCCC}
			\toprule
			& & & \multicolumn{3}{c}{\mysquare[blue] Car} & \multicolumn{3}{c}{\mysquare[red] Pedestrian} & \\
			\cmidrule(lr){4-6} \cmidrule(lr){7-9}
			Config & Training domain & + Ray-drop prior                                      & Precision     & Recall        & IoU           & Precision     & Recall        & IoU           & mIoU          \\
			\midrule
			A      & Simulation      &                                                       & 54.2          & 1.1           & 1.1           & 27.7          & 2.5           & 2.4           & 1.7           \\
			B      & Simulation      & + Global frequency                                    & 66.2          & 77.0          & 55.2          & 29.9          & 61.0          & 25.1          & 40.2          \\
			C      & Simulation      & + Pixel-wise frequency~\cite{wu2019squeezesegv2}      & 72.9          & 75.5          & 59.0          & 26.1          & 62.0          & 22.5          & 40.7          \\
			D      & Simulation      & + Auto-decoding w/ DUSty~\cite{nakashima2021learning} & 72.0          & 76.8          & 59.1          & \textbf{34.5} & 59.6          & \textbf{28.0} & 43.5          \\
			E      & Simulation      & + \textbf{Auto-decoding w/ ours}                      & \textbf{74.8} & \textbf{87.0} & \textbf{67.3} & 28.8          & \textbf{67.1} & 25.2          & \textbf{46.3} \\
			\midrule
			F      & Real            &                                                       & 78.7          & 86.5          & 70.1          & 66.5          & 18.0          & 16.5          & 43.3          \\
			\bottomrule
		\end{tabularx}
	\end{threeparttable}
\end{table*}

\begin{figure*}[tb]
	\footnotesize
	\begin{tabularx}{\hsize}{CCCCC}
		Input depth & Ground truth & Config-A & Config-C~\cite{wu2019squeezesegv2} & Config-E (\textbf{ours}) 
	\end{tabularx} \\
	\includegraphics[width=\hsize]{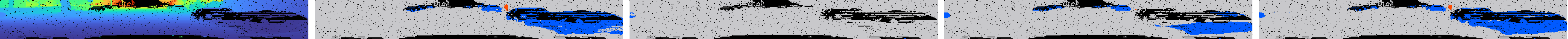} \\
	\includegraphics[width=\hsize]{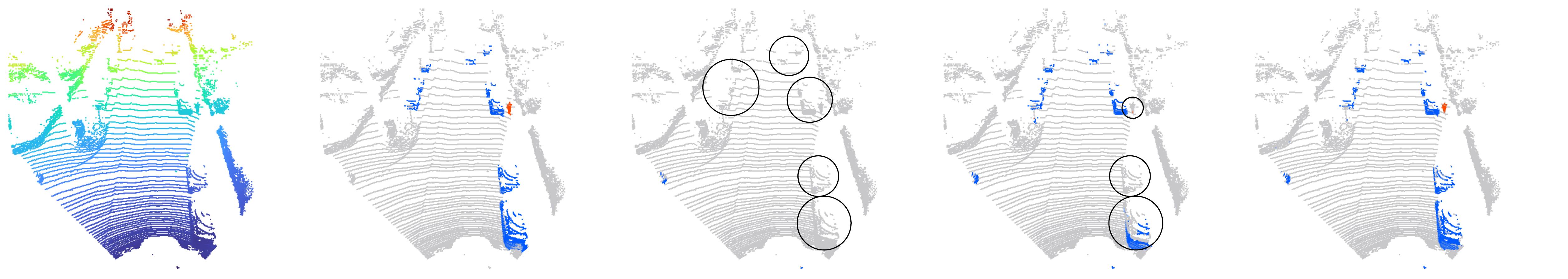} \\
	\begin{tabularx}{\hsize}{CCCCC}
		\includegraphics[height=4mm]{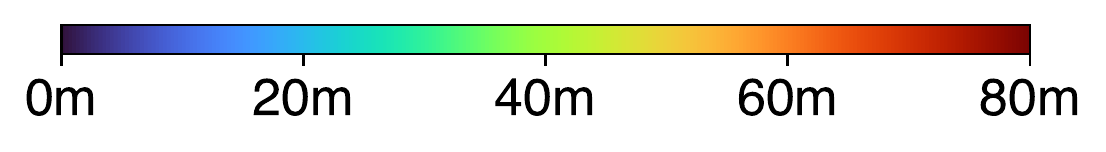} & \includegraphics[height=4mm]{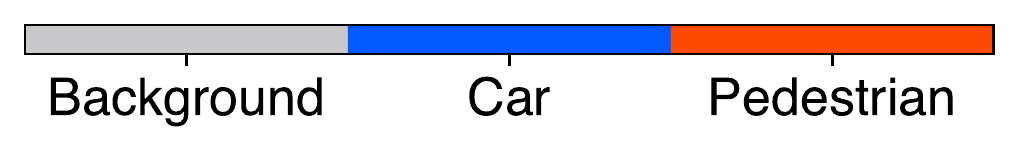} &   &   &   
	\end{tabularx}
	\caption{Qualitative comparison of Sim2Real semantic segmentation results. We can see that ours (config E) reduced the false negative on the \textcolor{blue}{\textit{car}} regions, as supported by the recall improvement in Table~\ref{tab:sim2real_noise}.}
	\label{fig:semseg_results_3d}
\end{figure*}

\subsection{Sim2Real semantic segmentation}
\label{sec:sim2real}

Model-based LiDAR simulators~\cite{wu2018squeezeseg,Dosovitskiy17} can produce a large amount of annotated training data, while there is an appearance gap against the real domain since the ray-drop noises are ignored or approximated.
Some studies address the issue by leveraging the ray-drop frequency~\cite{wu2018squeezeseg} or learning inference networks~\cite{zhao2021epointda,manivasagam2020lidarsim}.
As addressed in Section~\ref{sec:method:inference}, our auto-decoding process can also generate a ray-drop probability map $x_n$ by reconstructing valid points in a given range image $\hat{x}$.
This motivates us to reproduce the pseudo ray-drop noises on the simulated range images.
In this sections, we demonstrate the effectiveness of our method on the Sim2Real semantic segmentation.

\noindent
\textbf{Datasets.}
We follow the experiment protocol by Wu~\textit{et al.}~\cite{wu2019squeezesegv2} where a segmentation model is trained on the GTA-LiDAR dataset~\cite{wu2019squeezesegv2} and evaluated on the $90\degree$ frontal subset~\cite{wu2019squeezesegv2} of the KITTI dataset~\cite{geiger2013vision}, hereinafter called KITTI-frontal.
GTA-LiDAR is composed of 120k in-game LiDAR range images annotated with pixel-wise labels for \textit{car} and \textit{pedestrian} classes.
KITTI-frontal is composed of 10k real range images subsampled from KITTI, which is also annotated for the same classes.
KITTI-frontal contains 8,057 images for training and 2,791 images for testing.

\noindent
\textbf{Our approach.}
Before training the segmentation model, we perform the auto-decoding on each sample of GTA-LiDAR and obtain the corresponding ray-drop probability map $x_n$.
At training phase, we sample Bernoulli noises from $x_n$ and render the ray-drop noises on the fly.

\begin{table*}[tb]
	\centering
	\footnotesize
	\begin{threeparttable}
		\caption{Sim2Real performance in comparison with state-of-the-art domain adaptation (DA) methods. Following the prior work, we report precision (\%, $\uparrow$), recall (\%, $\uparrow$), and IoU (\%, $\uparrow$) for each class. The scores of DAN~\cite{long2015learning}, CORAL~\cite{sun2016deep}, HoMM~\cite{chen2020homm}, ADDA~\cite{tzeng2017adversarial}, and CyCADA~\cite{hoffman2018cycada} are from the report by Zhao \textit{et al.}~\cite{zhao2021epointda}.}
		\label{tab:sim2real_sota}
		\begin{tabularx}{\hsize}{l cc cccc CCC CCC C}
			\toprule
			& \multicolumn{2}{c}{DA$^\dag$} & \multicolumn{4}{c}{Input modality$^\ddag$} & \multicolumn{3}{c}{\mysquare[blue] Car} & \multicolumn{3}{c}{\mysquare[red] Pedestrian} & \\
			\cmidrule(lr){2-3} \cmidrule(lr){4-7} \cmidrule(lr){8-10} \cmidrule(lr){11-13}
			Method                                      & \texttt{D} & \texttt{F} & \texttt{C} & \texttt{R} & \texttt{I} & \texttt{M} & Precision     & Recall        & IoU           & Precision                             & Recall        & IoU           & mIoU          \\
			\midrule
			SqueezeSegV2~\cite{wu2019squeezesegv2}$^\S$ & \checkmark & \checkmark & \checkmark & \checkmark & \checkmark & \checkmark & --            & --            & 57.4          & --                                    & --            & 23.5          & 40.5          \\
			DAN~\cite{long2015learning}                 &            & \checkmark & \checkmark &            &            &            & 56.3          & 76.4          & 47.8          & 20.8                                  & \textbf{68.9} & 19.0          & 33.4          \\
			CORAL~\cite{sun2016deep}                    &            & \checkmark & \checkmark &            &            &            & 56.5          & 82.1          & 50.2          & 26.0                                  & 50.3          & 20.7          & 35.5          \\
			HoMM~\cite{chen2020homm}                    &            & \checkmark & \checkmark &            &            &            & 59.4          & 85.2          & 53.9          & 26.2                                  & 66.8          & 23.2          & 38.6          \\
			ADDA~\cite{tzeng2017adversarial}            &            & \checkmark & \checkmark &            &            &            & 56.7          & 83.5          & 50.7          & 24.7                                  & 58.5          & 21.0          & 35.9          \\
			CyCADA~\cite{hoffman2018cycada}             & \checkmark & \checkmark & \checkmark &            &            &            & 40.9          & 72.1          & 35.3          & 17.8                                  & 52.4          & 15.3          & 25.3          \\
			ePointDA~\cite{zhao2021epointda}$^\S$       & \checkmark & \checkmark & \checkmark &            &            &            & 73.4          & 81.9          & 63.4          & \textbf{29.4}                         & 56.0          & 23.9          & 43.7          \\
			ePointDA~\cite{zhao2021epointda}            & \checkmark & \checkmark & \checkmark &            &            &            & \textbf{75.2} & 84.7          & 66.2          & 28.7                                  & 65.2          & 24.8          & 45.5          \\
			\textbf{Ours (config-E)}$^\S$               & \checkmark &            & \checkmark & \checkmark &            &            & 74.8          & \textbf{87.0} & \textbf{67.3} & 28.8                                  & 67.1          & \textbf{25.2} & \textbf{46.3} \\
			\bottomrule
		\end{tabularx}
		\begin{tablenotes}
			\footnotesize
			\item $\dag$~Category of domain adaptation (DA): \texttt{D}: data-level DA and/or \texttt{F}: feature-level DA.
			\item $\ddag$~\texttt{C}: the Cartesian coordinates, \texttt{R}: depth, \texttt{I}: estimated intensity~\cite{wu2019squeezesegv2}, \texttt{M}: a binary mask indicating either measured or missing points.
			\item $\S$~The same model architecture (SqueezeSegV2~\cite{wu2019squeezesegv2}), but the different DA approach and input modality.
		\end{tablenotes}
	\end{threeparttable}
\end{table*}

\noindent
\textbf{Baselines.}
Our experiments are composed of two parts.
The first experiment compares five approaches with different ray-drop priors, as listed in Table~\ref{tab:sim2real_noise}.
Config-A is a baseline without rendering ray-drop noises.
In config-B, we sample Bernoulli noises from global frequency computed from all pixels in KITTI-frontal.
Config-C is the approach used by Wu~\textit{et al.}~\cite{wu2019squeezesegv2}, where the noises are sampled from pixel-wise frequency of KITTI-frontal.
Finally, config-D and config-E are the GAN-based auto-decoding approach.
Config-D uses DUSty~\cite{nakashima2021learning}, while config-E uses our proposed GAN.
Both models are pretrained on KITTI in Section~\ref{sec:fidelity_and_diversity}.
For comparison, we also provide the oracle results trained on KITTI-frontal (config-F).
We use SqueezeSegV2~\cite{wu2019squeezesegv2} for the architecture of semantic segmentation.
To demonstrate the exclusive effect of noise rendering, we do not use any other adaptation techniques used in SqueezeSegV2, such as learned intensity rendering, geodesic correlation alignment, and progressive domain calibration.
In the second experiment, we compare our model (config-E) with state-of-the-art domain adaptation methods: DAN~\cite{long2015learning}, CORAL~\cite{sun2016deep}, HoMM~\cite{chen2020homm}, ADDA~\cite{tzeng2017adversarial}, CyCADA~\cite{hoffman2018cycada}, and ePointDA~\cite{zhao2021epointda}.
ePointDA is a CycleGAN-based method and closely related to us in that simulating ray-drop noises on range images.

\begin{figure}[tb]
	\footnotesize
	\centering
	\setlength\tabcolsep{3pt}
	\begin{tabularx}{\hsize}{lCC}
		                                                              &                                                               
		{
		\begin{tabularx}{\hsize}{CC}
		Target $\hat{x}$                                              & $x_G$                                                         \\
		\includegraphics[width=\hsize]{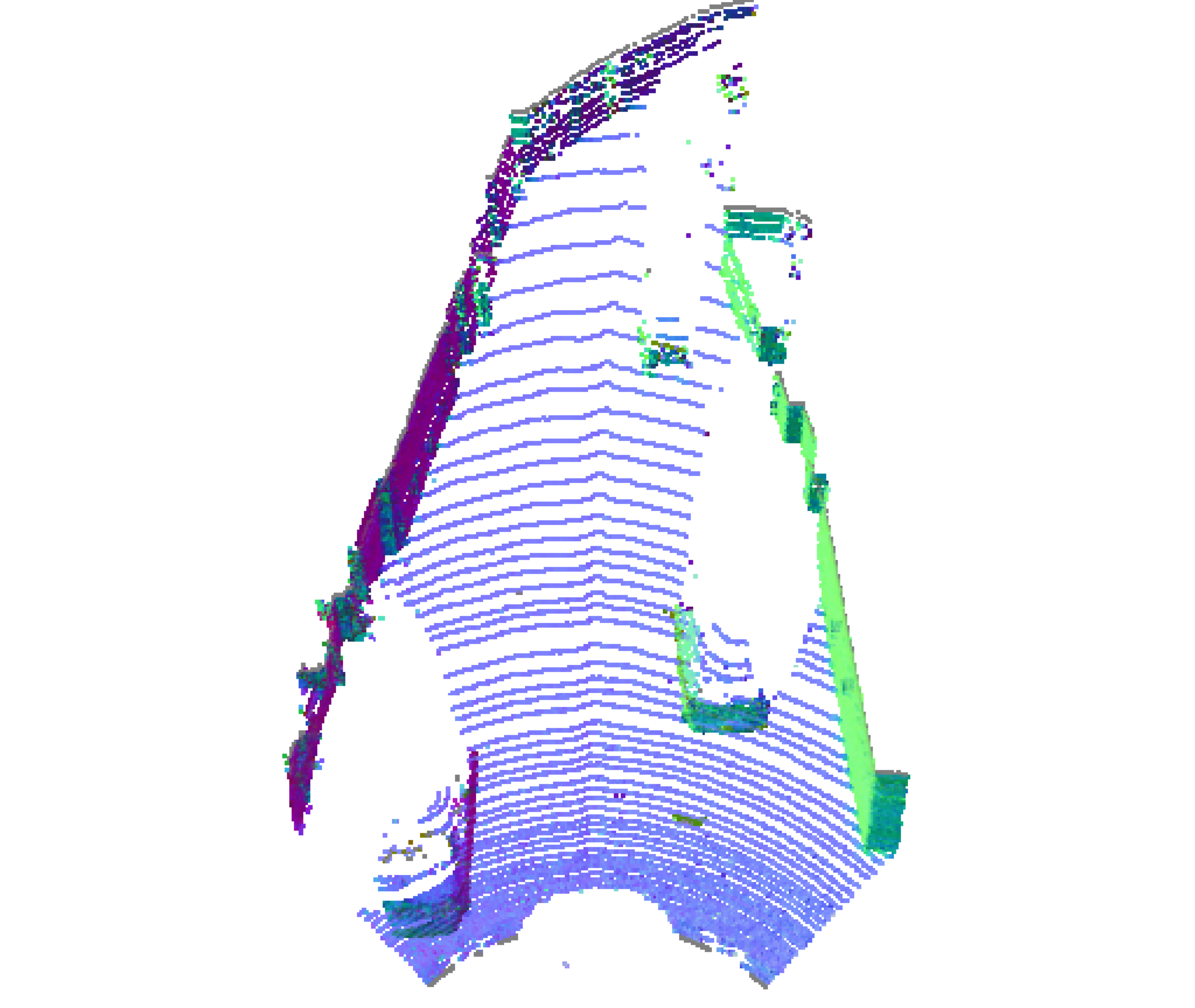} & \includegraphics[width=\hsize]{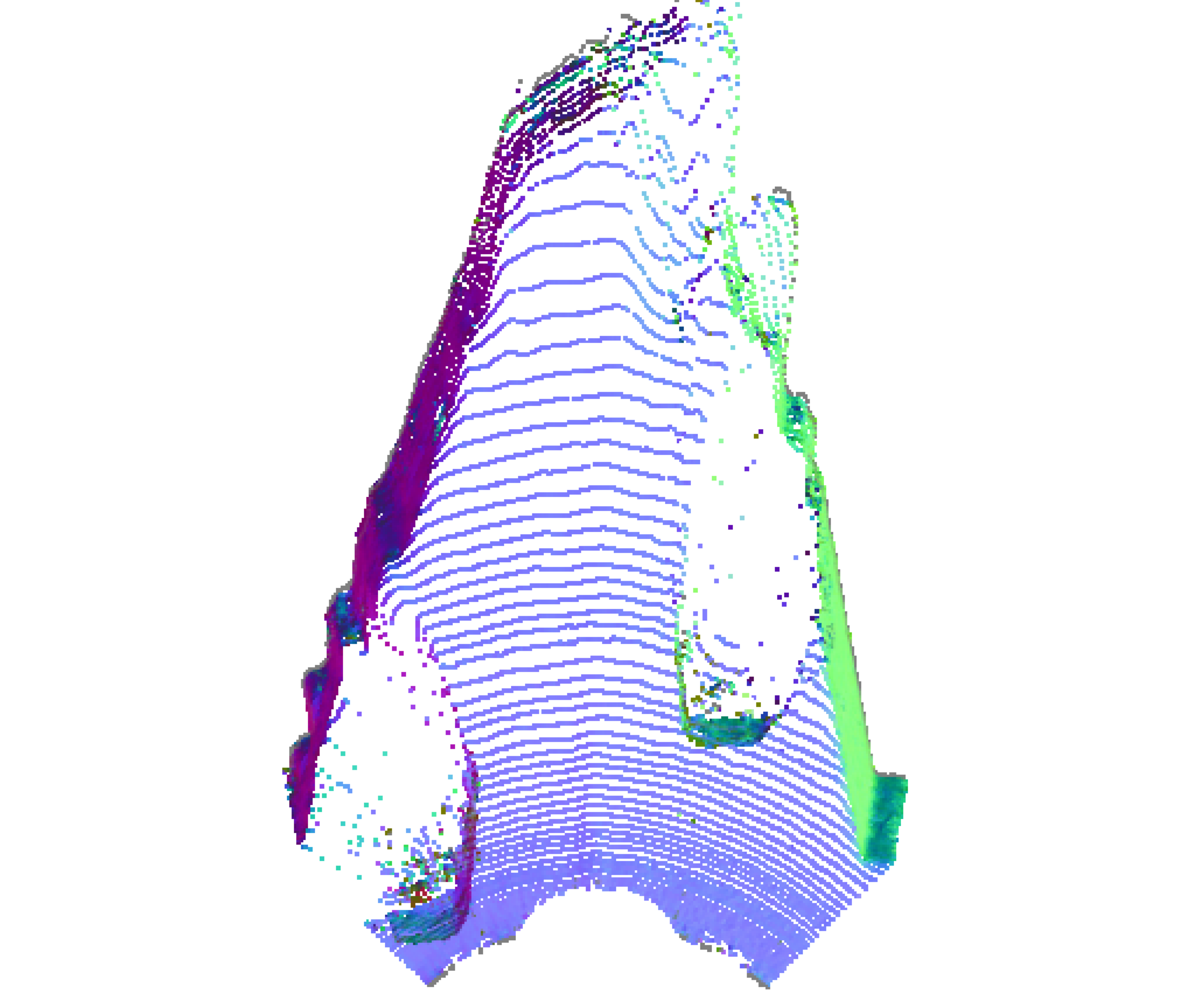} 
	\end{tabularx}
	}
	&
	{
		\begin{tabularx}{\hsize}{CC}
			Target $\hat{x}$                                            & $x_G$                                                       \\
			\includegraphics[width=\hsize]{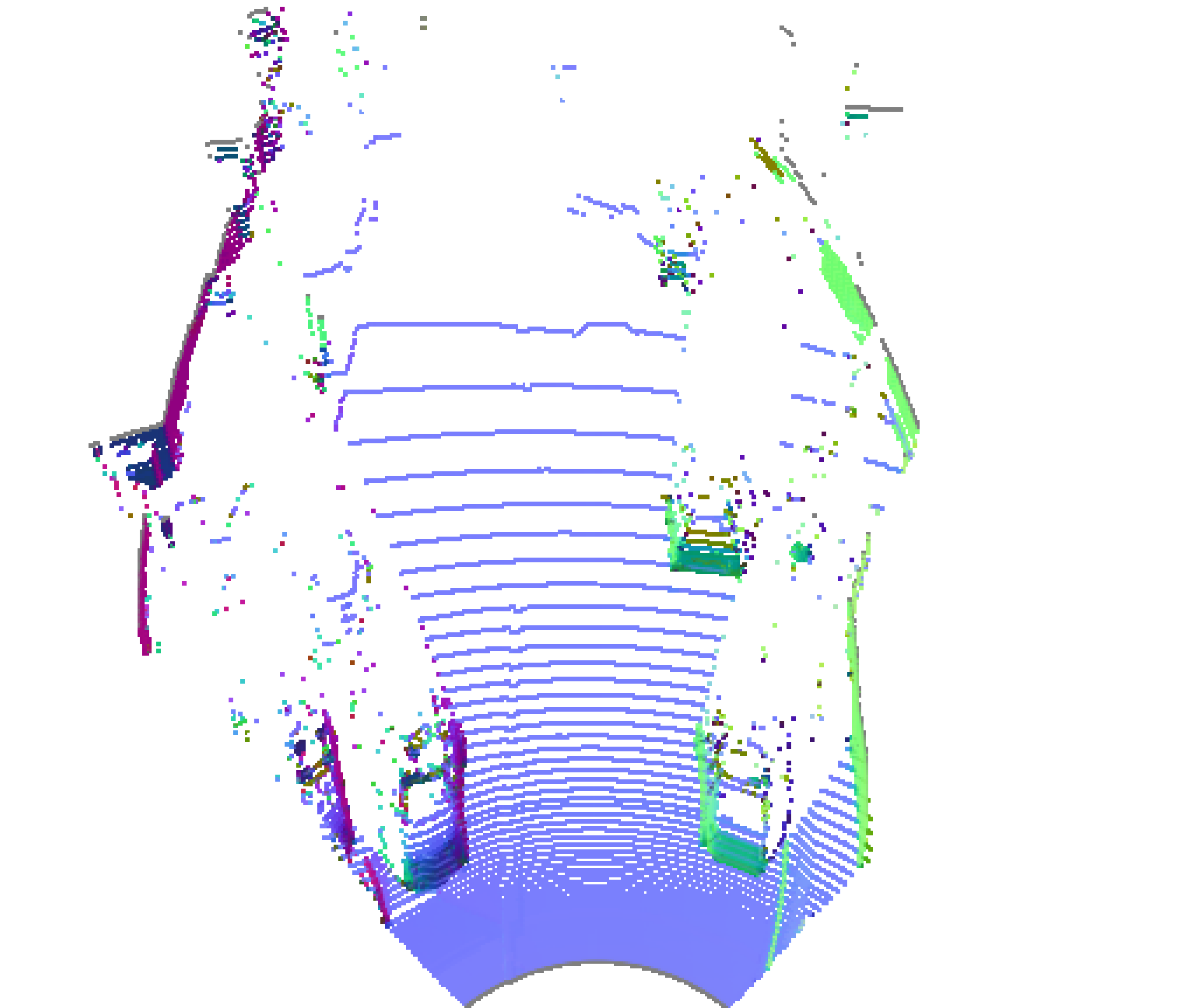} & \includegraphics[width=\hsize]{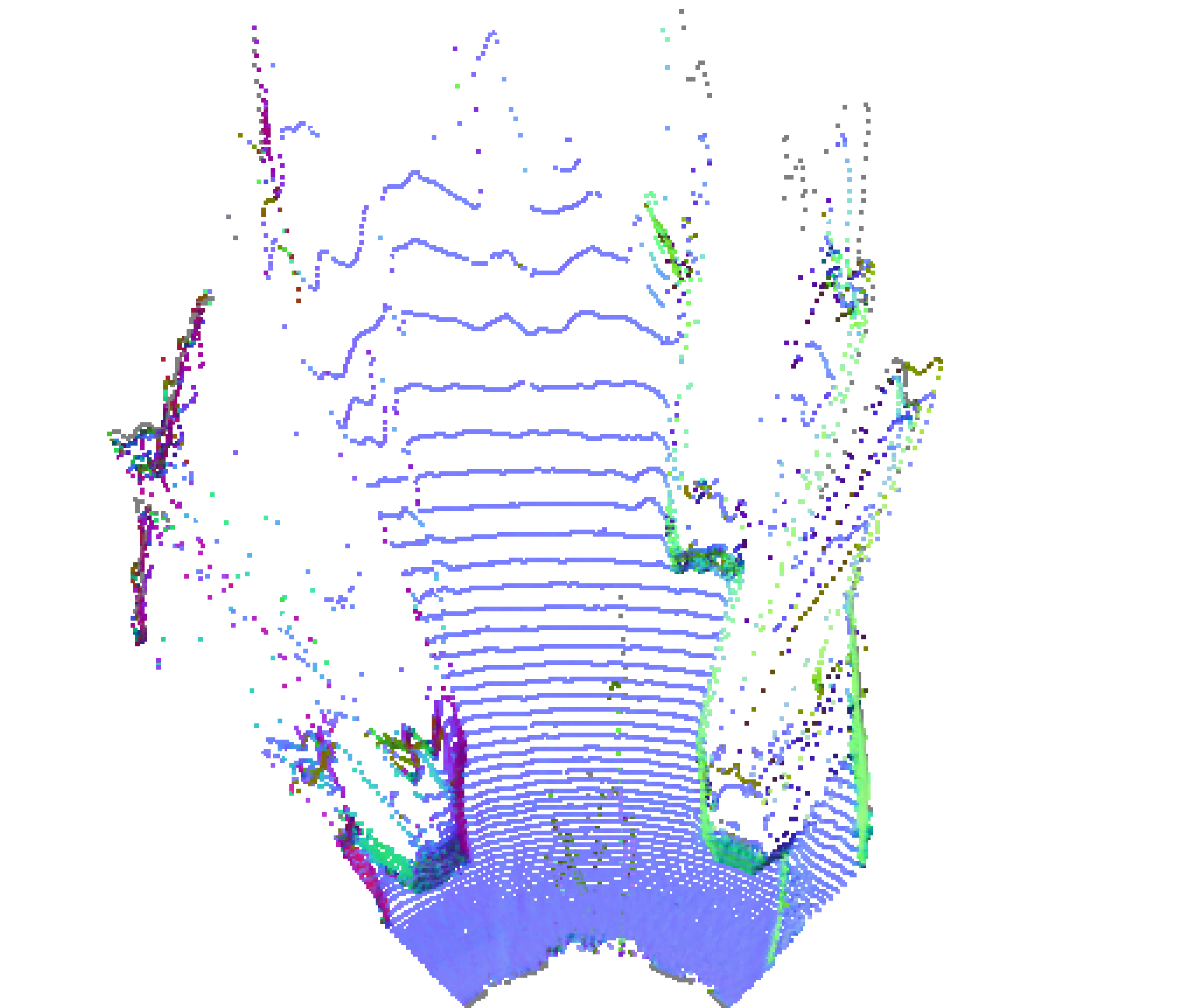} 
		\end{tabularx}
	}
	\\
	$\hat{x}$ & \includegraphics[width=\hsize]{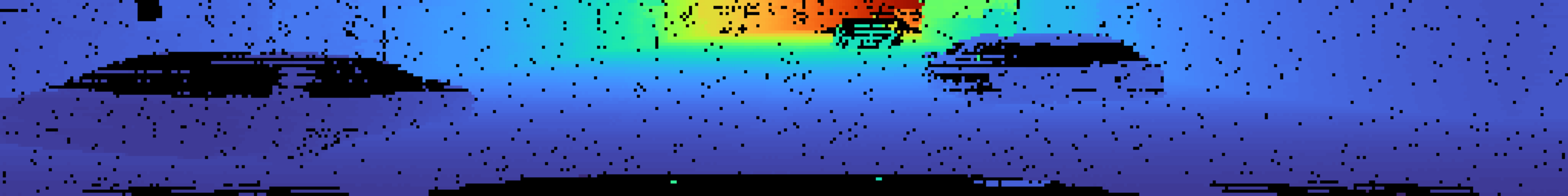} & \includegraphics[width=\hsize]{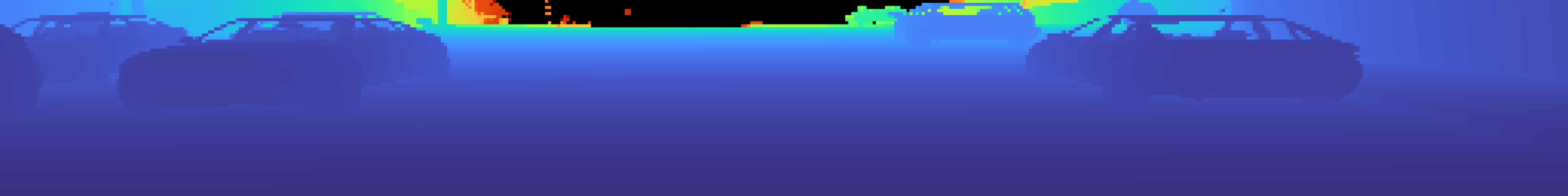} \\
	$x_d$ & \includegraphics[width=\hsize]{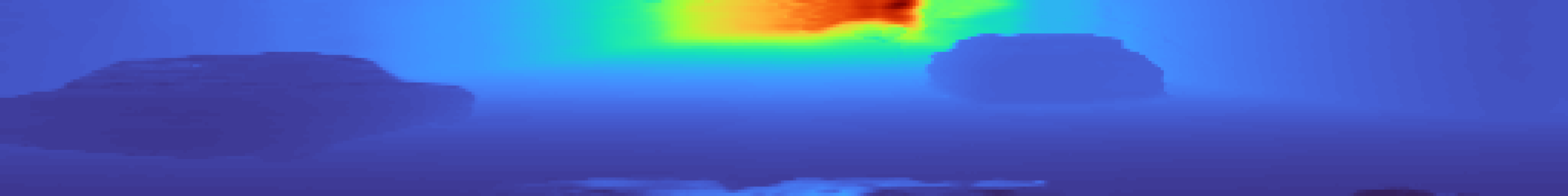} & \includegraphics[width=\hsize]{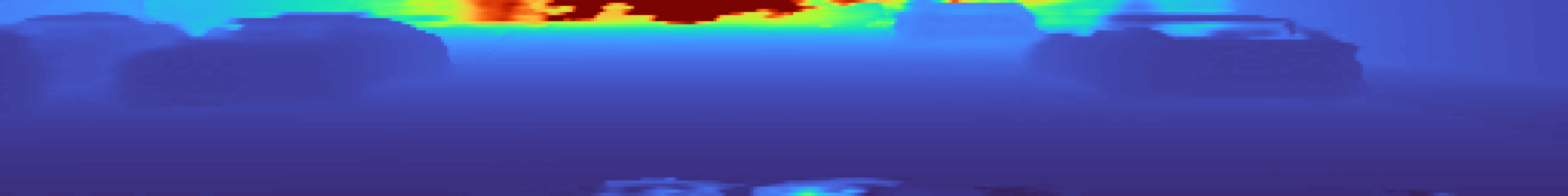} \\
	$x_G$ & \includegraphics[width=\hsize]{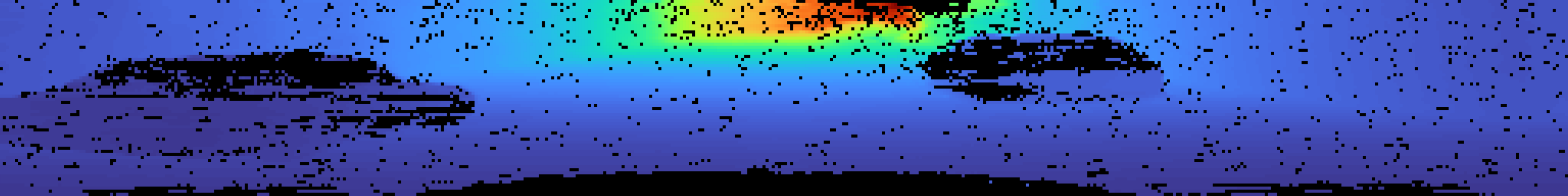} & \includegraphics[width=\hsize]{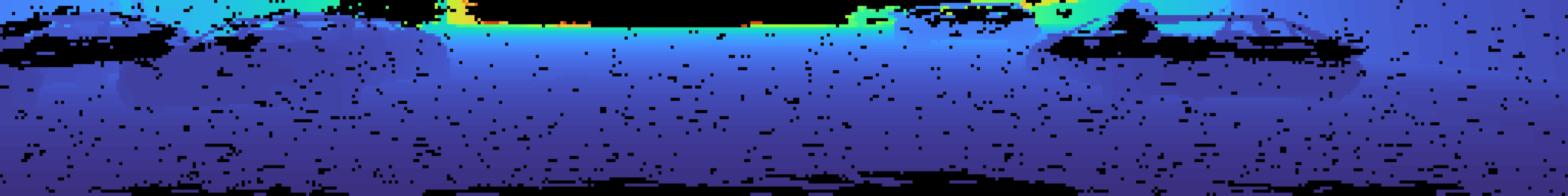} \\
	& KITTI-frontal & GTA-LiDAR \\
	& (real domain) & (simulation domain) \\
	\end{tabularx}
	\vspace{0.5mm}
	\caption{Auto-decoding examples by our method. We show the targets $\hat{x}$, the intermediate outputs $x_d$, and the final outputs $x_G$.}
	\label{fig:sim2real_autodecoding}
\end{figure}

\begin{figure}[tb]
	\footnotesize
	\centering
	\setlength\tabcolsep{4pt}
	\begin{tabularx}{\hsize}{rLC}
		A: & Simulated range image~\cite{wu2019squeezesegv2}        & \includegraphics[width=\hsize]{figs/sim2real/gta_A_img.pdf}   \\
		   & \mysquare[blue] Car / \mysquare[red] pedestrian labels & \includegraphics[width=\hsize]{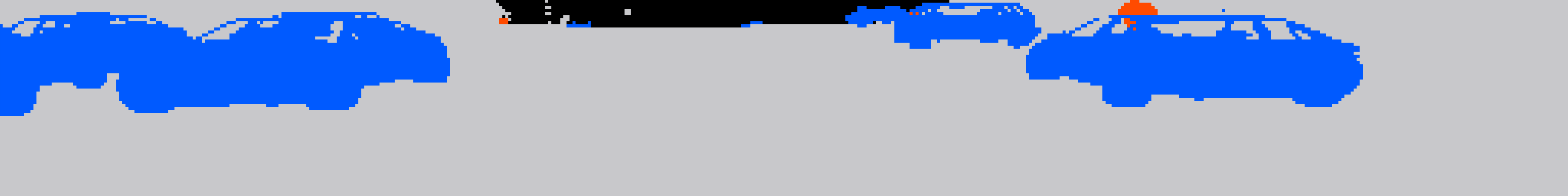} \\
		B: & Global frequency                                       & \includegraphics[width=\hsize]{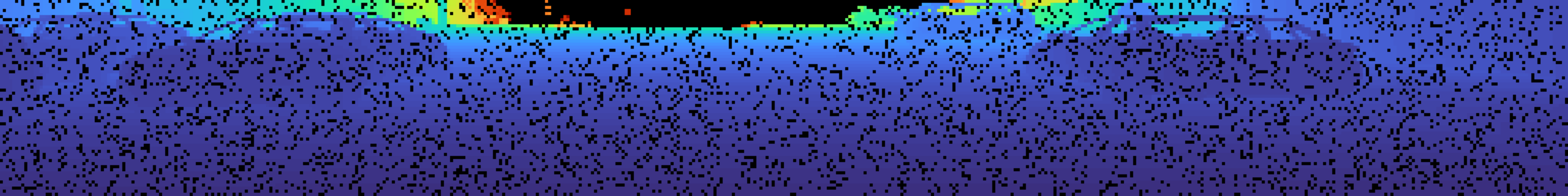}   \\
		C: & Pixel-wise frequency~\cite{wu2018squeezeseg}           & \includegraphics[width=\hsize]{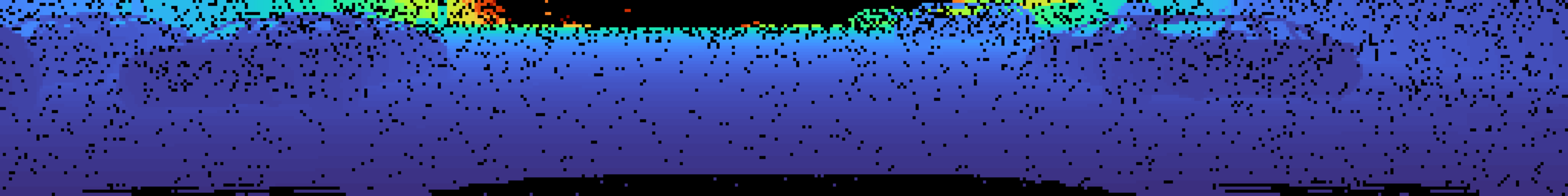}   \\
		D: & Auto-decoding w/ DUSty~\cite{nakashima2021learning}    & \includegraphics[width=\hsize]{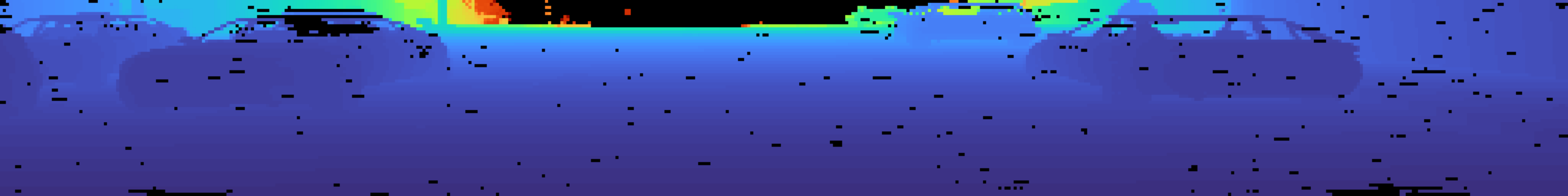}   \\
		E: & \textbf{Auto-decoding w/ ours}                         & \includegraphics[width=\hsize]{figs/sim2real/gta_E_img.pdf}   \\
		   &                                                        & \includegraphics[height=5mm]{figs/colorbars/depth.pdf}        \\
	\end{tabularx}
	\vspace{-2mm}
	\caption{Qualitative comparison of noise rendering methods on GTA-LiDAR~\cite{wu2019squeezesegv2}.}
	\label{fig:sim2real_examples}
\end{figure}

\noindent
\textbf{Results.}
Table~\ref{tab:sim2real_noise} reports intersection-over-union (IoU) for each class and the mean score (mIoU).
Fig.~\ref{fig:semseg_results_3d} provides visual comparison of config-A without noises, config-C~\cite{wu2019squeezesegv2}, and our config-E.
Although SqueezeSegV2 is already designed to reduce the ray-drop sensitivity, the performance is extremely low without noise rendering (config-A).
Surprisingly, even simple rendering with the global frequency (config-B) boosted all the metrics and its spatial extension by config-C~\cite{wu2018squeezeseg} brought subtle improvement.
GAN-based approaches (config-D and config-E) further improved the results.
In particular, our model (config-E) showed the best mIoU and surpassed the results on real domain (config-F).
Fig.~\ref{fig:sim2real_autodecoding} shows the auto-decoded examples and Fig.~\ref{fig:sim2real_examples} compares rendered noises.
We can see that our method successfully simulated instance-level ray-drops such as on the car body and ego-vehicle shadows.
In contrast, the results by the global frequency and the pixel-wise frequency are approximated.
Finally, Table~\ref{tab:sim2real_sota} compares our results (config-E) with state-of-the-art domain adaptation methods.
Despite not applying any domain adaptation techniques other than rendering noises, our model showed the best IoUs.

\section{Conclusion}

In this paper, we introduced a novel approach for learning data priors of 3D LiDAR data towards domain adaptation applications.
Our key idea is to represent LiDAR range images by a coordinate-based generative model and to learn clean data space through a pseudo-measurement model.
We designed our model based on state-of-the-art GANs and demonstrated its effectiveness in the LiDAR domain.
First, we evaluated the generation fidelity and diversity of sampled data. 
Our model showed superior results against the image-based and point-based baselines.
We also conducted a Sim2Real semantic segmentation using our learned ray-drop priors.
Our instance-level noise simulation brought significant improvement qualitatively and quantitatively and outperformed state-of-the-art methods.
The results revealed that rendering ray-drop noises is important to mitigate a gap between the real and simulation domains. 
We consider our sensor-agnostic scene representation has the potential for cross-dataset tasks.
Future work includes domain adaptation between different LiDARs and mixing accessible datasets for further training.

\ifwacvfinal
\section*{Acknowledgements}

This work was partially supported by a Grant-in-Aid for JSPS Fellows Grant Number JP19J12159, JSPS KAKENHI Grant Number JP20H00230, and JST [Moonshot R\&D] [Grant Number JPMJMS2032].
\fi

{\small
	\bibliographystyle{ieee_fullname}
	\bibliography{egbib}
}

\clearpage

\begin{appendices}
		
	\section{Overview}
		
	This supplementary material summarizes implementation details of our model architectures and experiments in Section~\ref{sec:implementation_details}, 
	detailed analysis of evaluation metrics in Section~\ref{sec:analysys_of_metrics}, 
	generated examples of our method and baselines in Section~\ref{sec:generated_examples},
	and Sim2Real semantic segmentation results in Section~\ref{sec:sim2real_examples}.
		
	\section{Implementation details}
	\label{sec:implementation_details}
		
	\subsection{Models}
		
	\begin{figure*}[t]
		\includegraphics[width=\hsize]{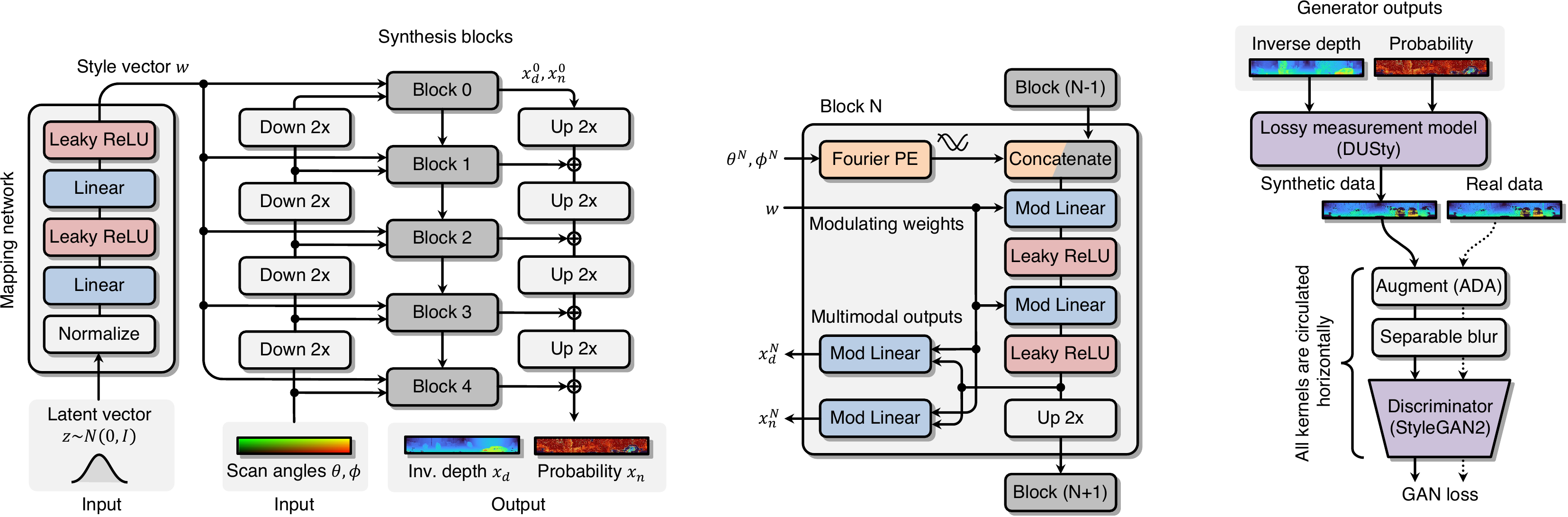}\\
		\begin{tabularx}{\hsize}{CCr}
			(a) An overview of our generator & (b) A detail of synthesis block & (c) Discriminator 
		\end{tabularx}
		\caption{Building blocks of our proposed GAN framework.}
		\label{fig:building_blocks}
	\end{figure*}
		
	Fig.~\ref{fig:building_blocks} shows an overview of our proposed GAN framework.
	We design the generator network based on INR-GAN~\cite{Skorokhodov_2021_CVPR}, which was proposed to generate natural images in coordinate-based representation.
		
	\noindent
	\textbf{Generator.}
	The generator is composed of a mapping network and synthesis blocks as shown in Fig.~\ref{fig:building_blocks}a.
	The mapping network transforms the latent space $z \sim N(0,I)$ into another representation, style space $w$, which modulates the weights of the synthesis blocks $\Omega$.
	The synthesis blocks represent the function which returns inverse depth $x_d$ and ray-drop probability $x_n$ given the specific angles $\Phi=(\theta, \phi)$.
	The outputs $x_d$ and $x_n$ are then converted to the final LiDAR image $x_G$ through the lossy measurement model.
	Each synthesis block encodes the angular inputs to high-dimensional space to represent spatial bias using Fourier features~\cite{tancik2020fourier}.
	Note that all operations in synthesis blocks are pixel-independent while the set of angles $\Phi$ is dowmsampled hierarchically to perform with a reasonable computational cost as proposed in INR-GAN.
		
	\noindent
	\textbf{Discriminator.}
	For the discriminator in Fig.~\ref{fig:building_blocks}c, we use the same setup of DUSty~\cite{nakashima2021learning} while replace the backbone with StyleGAN2~\cite{karras2020analyzing}.
	We applied the separable blur filter~\cite{kaneko2020noise} to the discriminator inputs and modify all the kernels with circular padding.
		
	\subsection{Training}
		
	We employed the adaptive discriminator augmentation (ADA)~\cite{karras2020training} for all the image-based methods: vanilla GAN, DUSty, and ours.
	The augmentation basically followed the original pipeline by Karras \textit{et al.}~\cite{karras2020training}, but disabled the steps of rotation and horizontal scaling that break the circular structure of range images.
	We also modified the integer/fractional translation into circulating behavior.
	We believe that it is required to explore the optimal augmentation set for LiDAR range images, while the tuning remains for future work.
		
	As the adversarial objective, we employed the non-saturating loss with a gradient penalty~\cite{karras2020analyzing}.
	The penalty coefficient was set to 1.
	All parameters were updated by Adam optimizer for 25M iterations with a learning rate of 0.002 and a batch size of 48. Training were performed on three NVIDIA RTX 3090 GPUs.
		
	\subsection{Computational cost of EMD}
		
\begin{figure}[t]
		\centering
		\includegraphics[width=0.9\hsize]{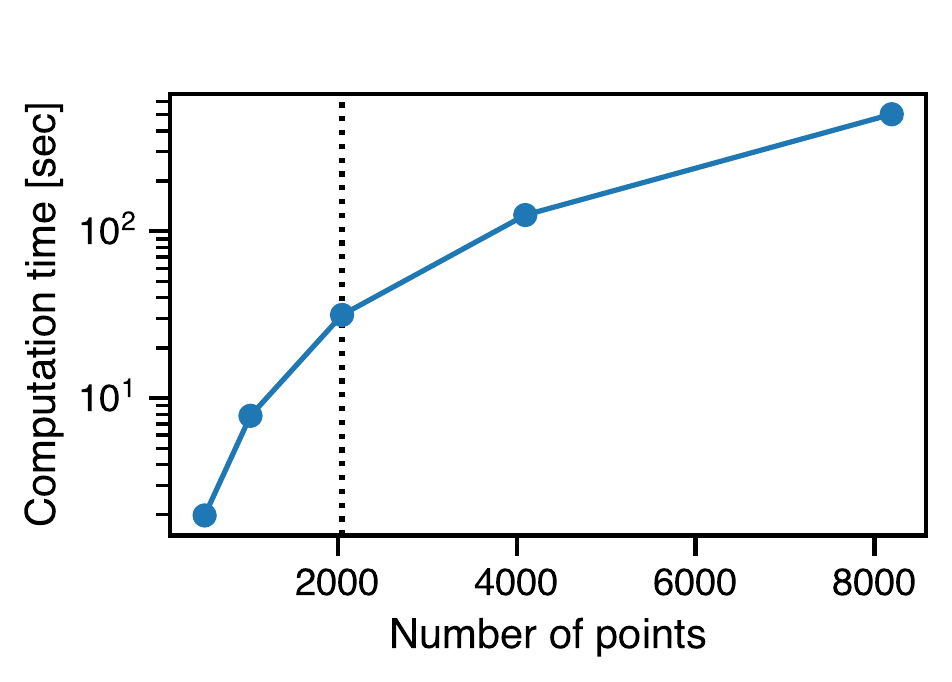}
		\caption{EMD computation time as a function of the number of points. The conventional number of point cloud tasks is 2048 (dotted line), while our task uses $64\times512=32,768$ points in full setting.}
		\label{fig:emd_computing_time}
	\end{figure}

	Earth mover's distance (EMD) is one of the metrics measuring the error between point clouds.
	Compared to the other metrics such as chamfer distance, EMD reflects the local details and the density distribution and is popular for the assessment of point clouds. 
	However, it is known that computing EMD has an $o(N^3)$ complexity where $N$ is the number of points in 3D point clouds~\cite{pele2009fast}. 
	This is problematic for our case using LiDAR point clouds, for instance, in training point-based models such as l-WGAN~\cite{achlioptas2018learning} and computing the standard evaluation metrics such as COV, MMD, and 1-NNA~\cite{yang2019pointflow}.
	In Fig.~\ref{fig:emd_computing_time}, we compute a pairwise distance of $M=10,000$ sets of $N$ points, where $N$ ranges from $2^9$ to $2^{13}$ with a batch size of 256, and show the computation time as a function of the number of points.
	Similar to Nakashima \textit{et al.}~\cite{nakashima2021learning}, we reduce the number of points to conventional 2048 by farthest point sampling in conducting experiments with point-based methods and evaluating the point-based metrics.
		
	\subsection{Inference}
		
	For the inference application, we use the style code $w$ instead of the latent code $z$ to gain reconstruction fidelity as demonstrated in the related studies~\cite{karras2020analyzing,Skorokhodov_2021_CVPR,anokhin2021image,karras2020training,roich2021pivotal}.
	We optimize the style code $w$ for 500 iterations in the first step (GAN inversion) and then optimize the generator weights $\Omega$ for another 500 iterations for the second step (pivotal tuning). We empirically set the learning rate for 0.05 and 0.0005 for the first and second steps, respectively.
		
	\section{Sanity check of evaluation metrics}
	\label{sec:analysys_of_metrics}

\begin{figure*}[t]
		\centering
		\begin{tabularx}{\hsize}{CC}
			\includegraphics[width=\hsize]{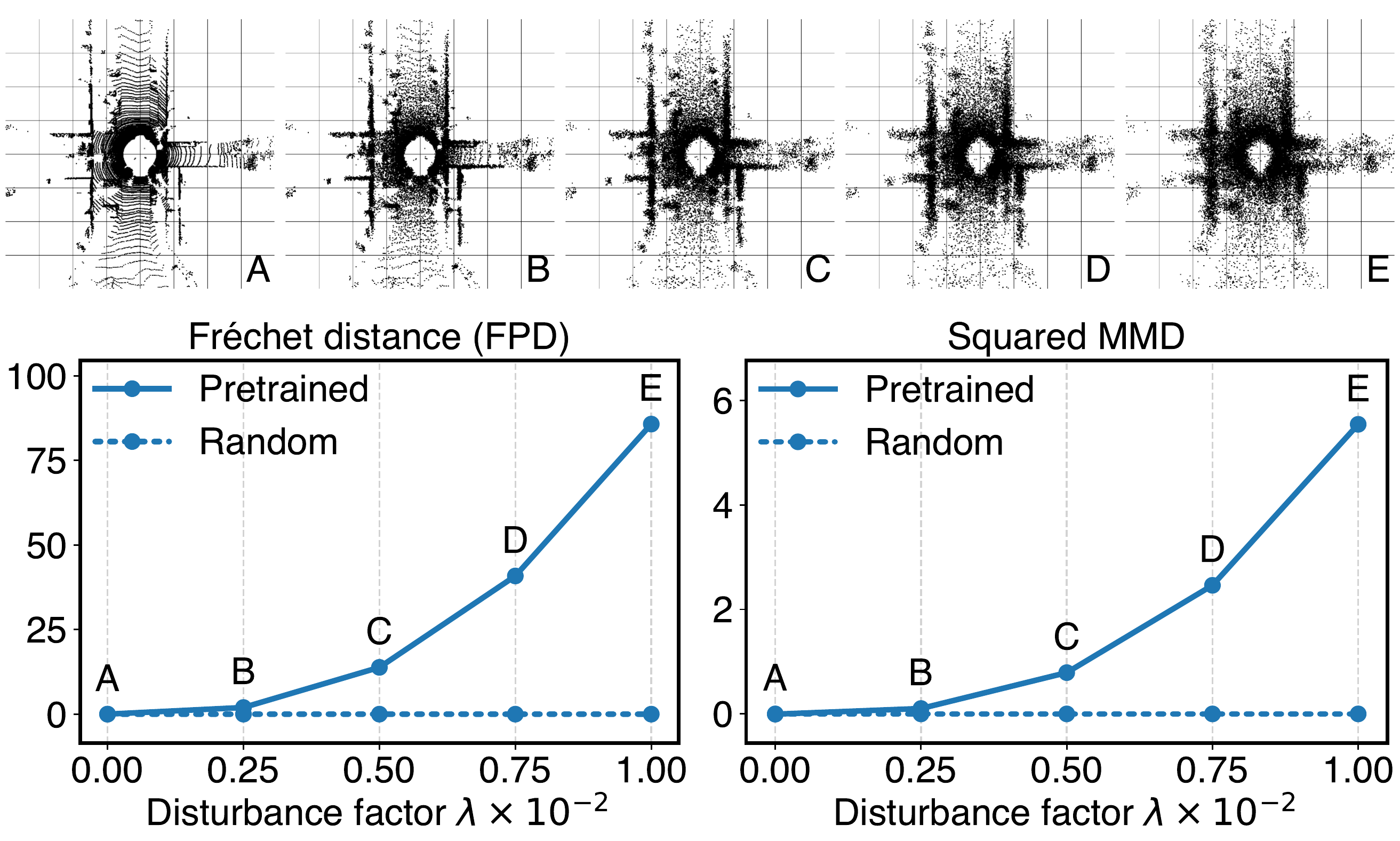}    & \includegraphics[width=\hsize]{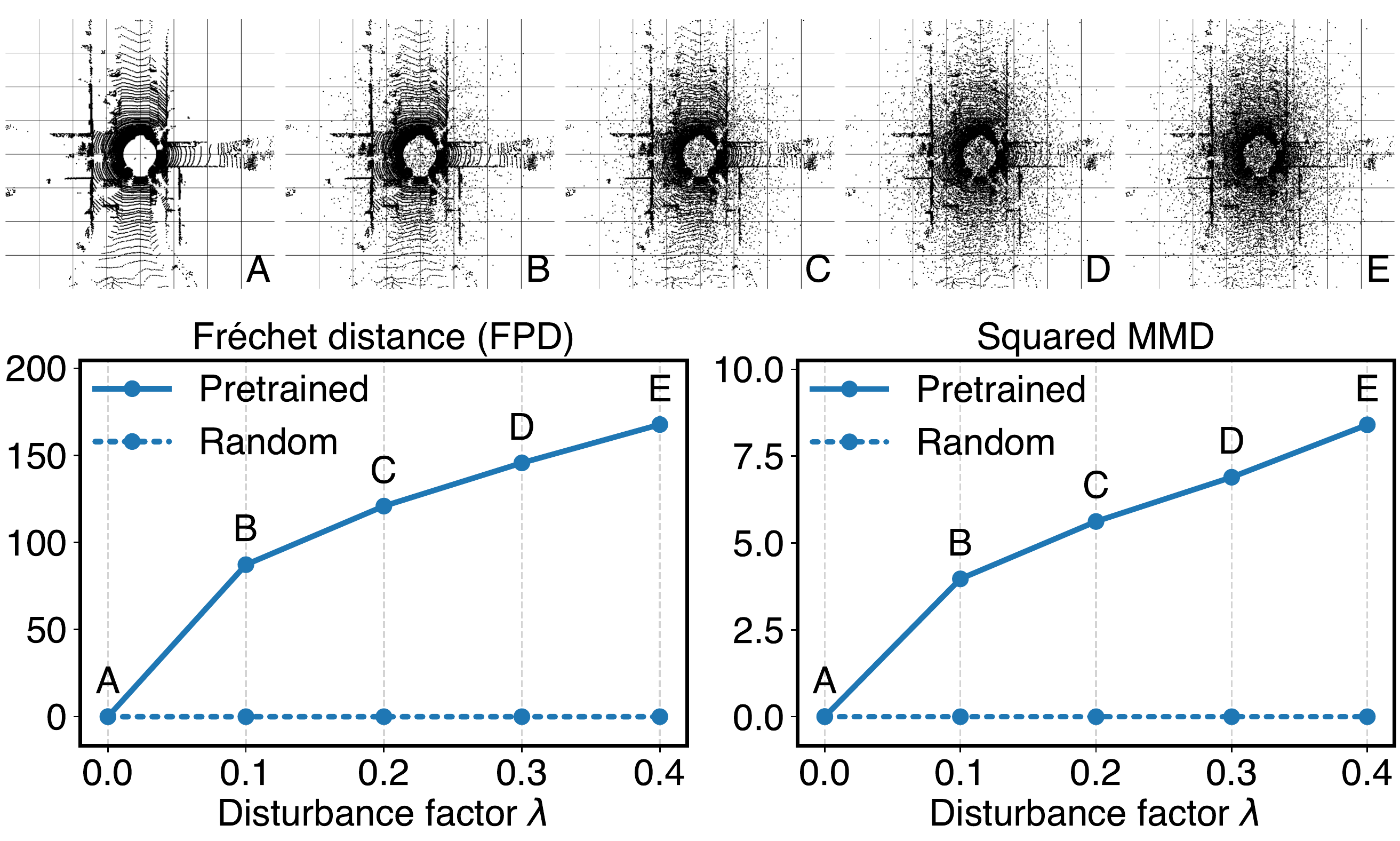}          \\
			\\
			(a) Additive Gaussian noises with a coefficient $\lambda$         & (b) Drop-in Gaussian noises for $\lambda \times100$ (\%) of points \\ 
			\\
			\\
			\includegraphics[width=\hsize]{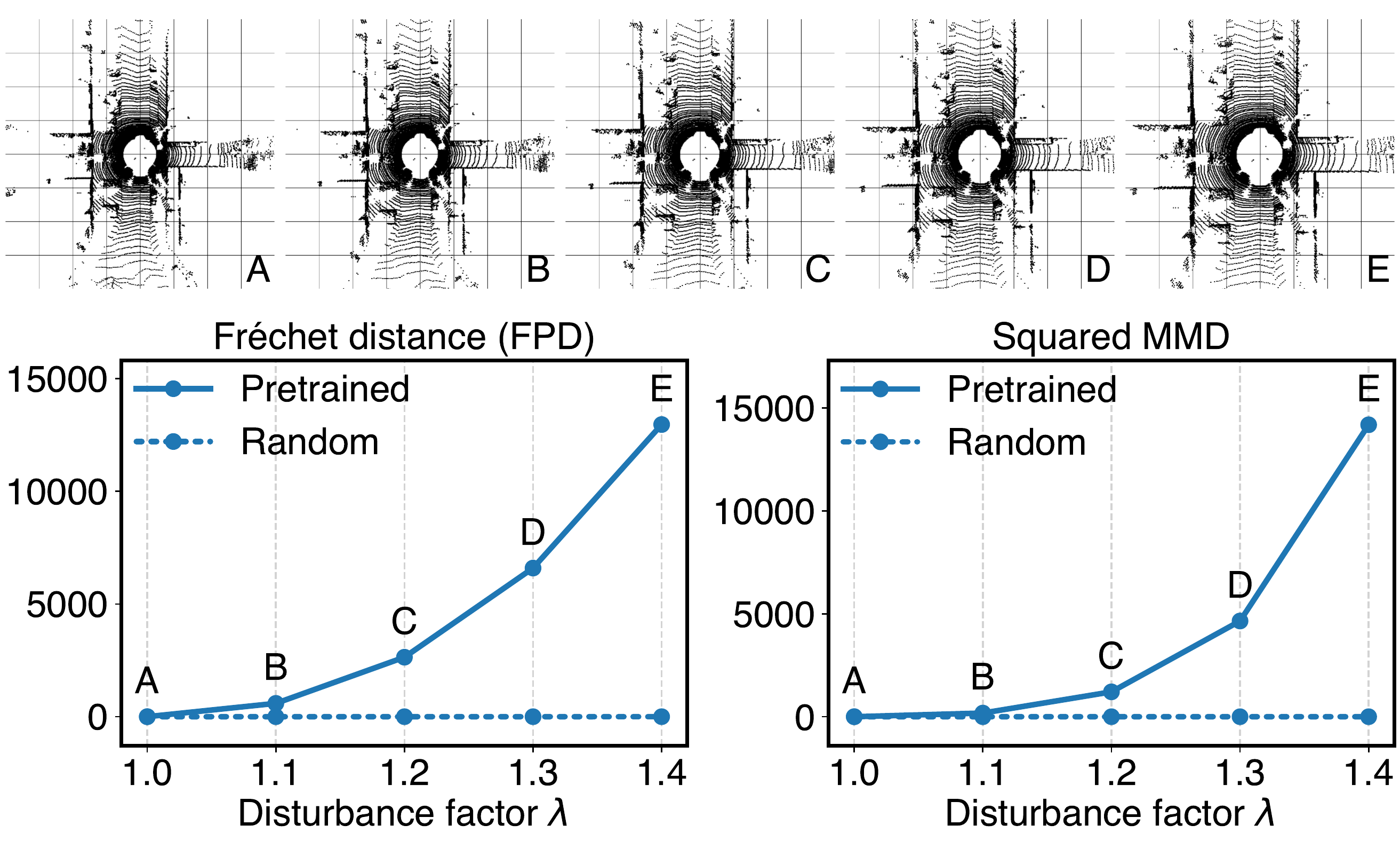}   & \includegraphics[width=\hsize]{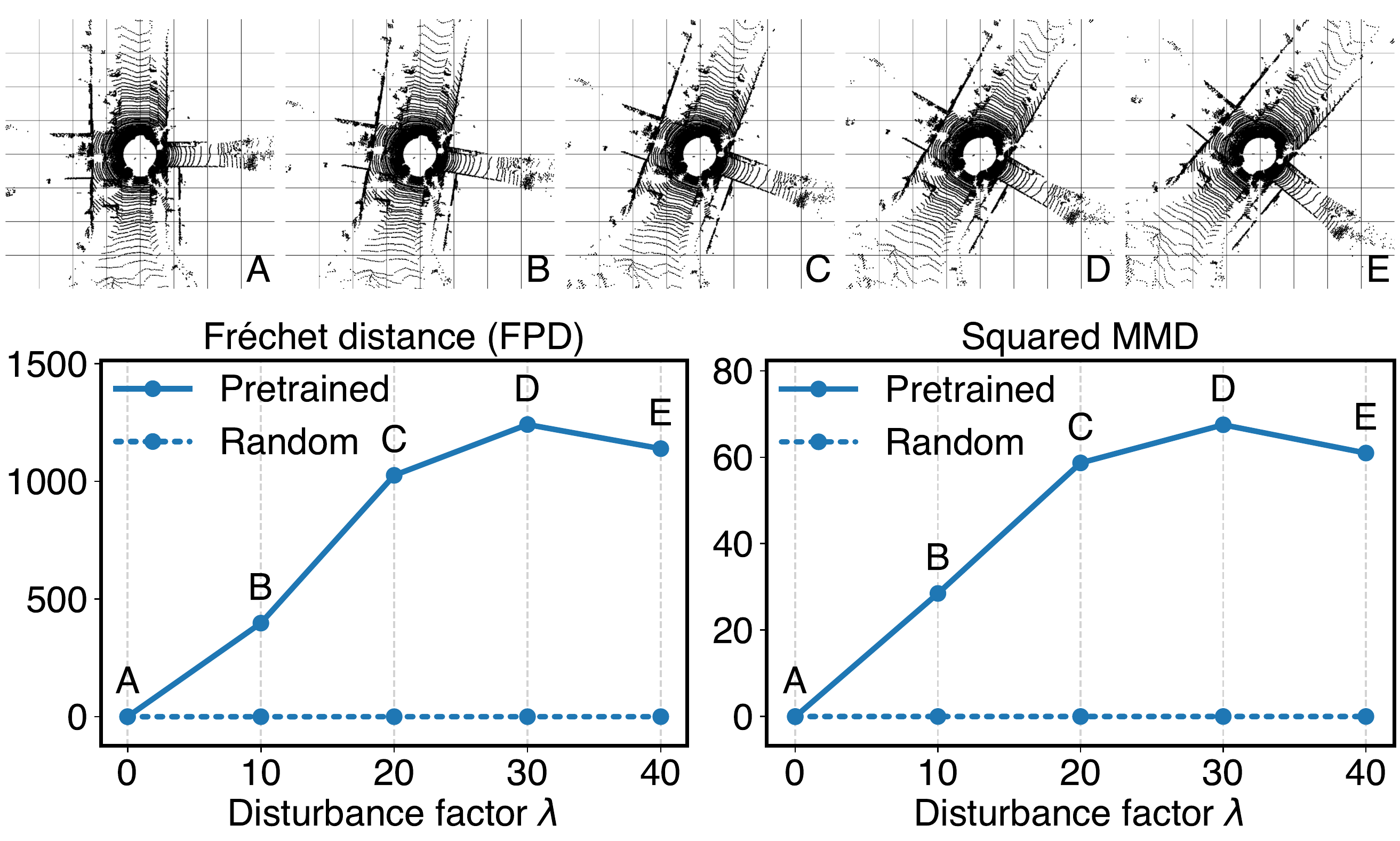}     \\
			\\
			(c) Inflating coordinates with a multiplicative factor $\lambda$  & (d) Clockwise yaw rotation with an angle $\lambda$ ($\degree$)     \\ 
			\\
			\\
			\includegraphics[width=\hsize]{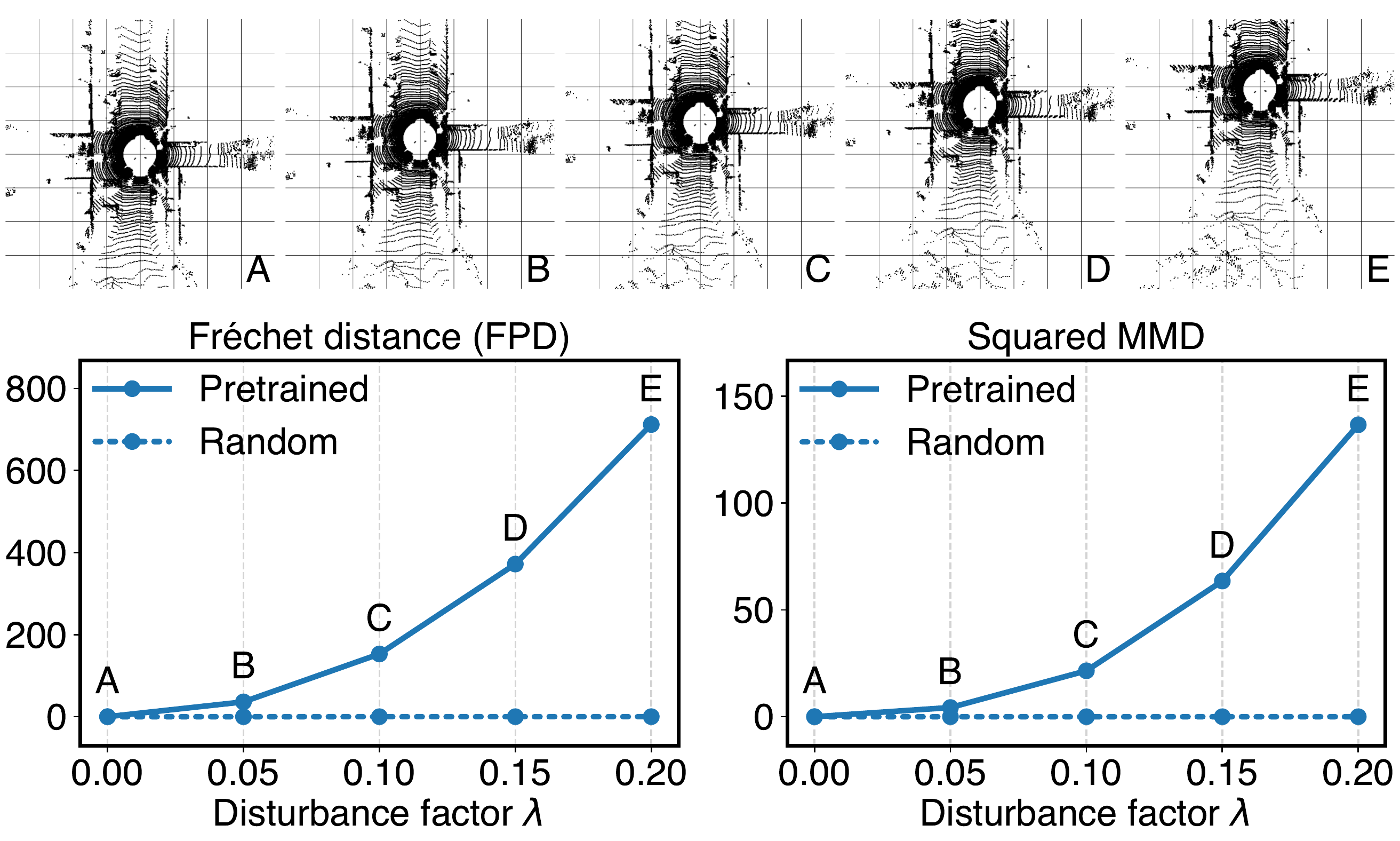} & \includegraphics[width=\hsize]{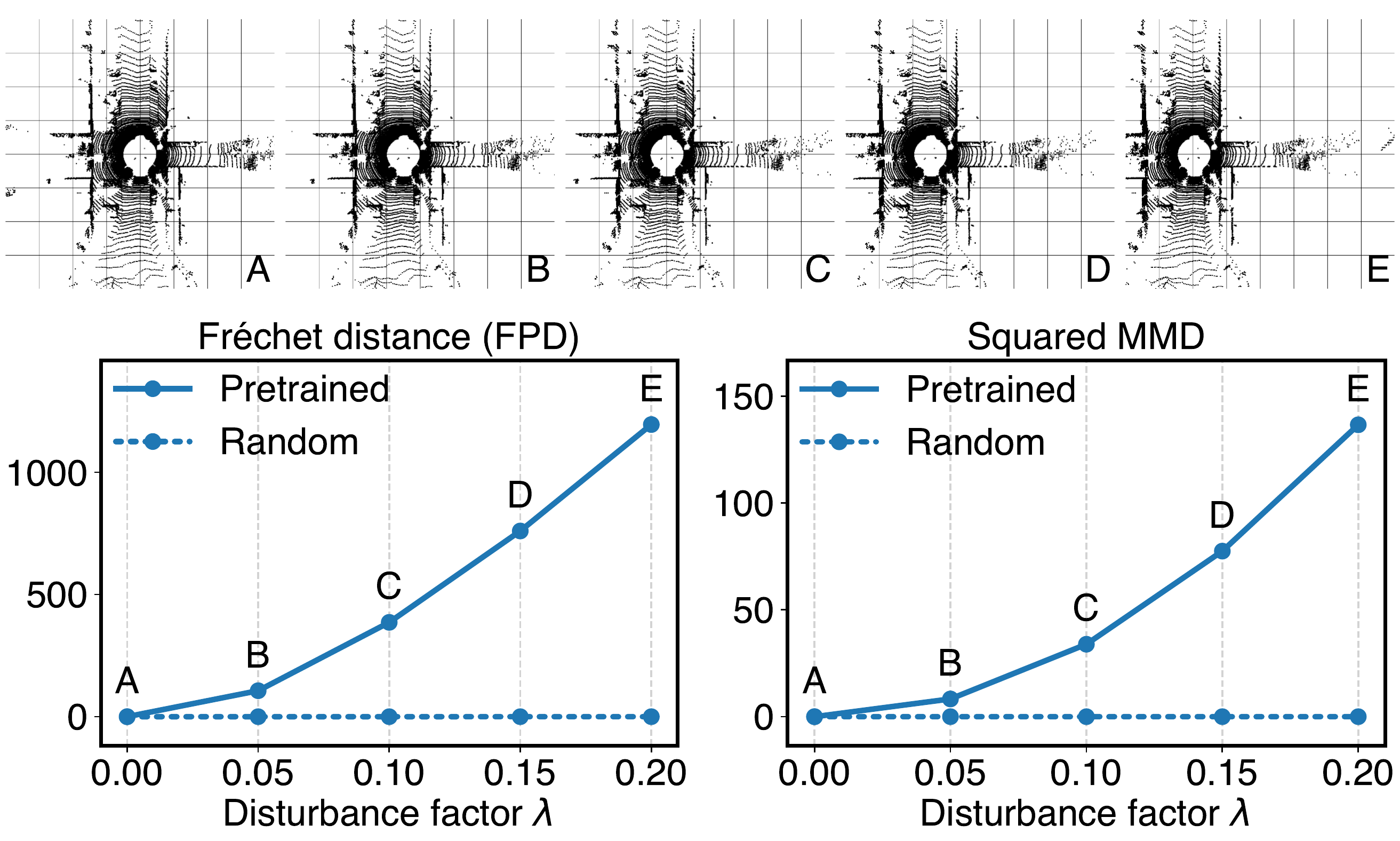}  \\
			\\
			(e) Translation in $x$ direction by $\lambda$                       & (f) Translation in $y$ direction by $\lambda$                      
		\end{tabularx}
		\caption{Disturbance sensitivity of four metrics: FPD~\cite{Shu_2019_ICCV} (Fréchet distance for point clouds) and squared maximum mean discrepancy (squared MMD)~\cite{binkowski2018demystifying}. We applied six types of disturbances to the KITTI point clouds with various strength (see A--E) and computed the metrics with the \textit{clean} original point clouds. All point clouds were encoded by PointNet~\cite{qi2017pointnet} with pre-trained or random weights.}
		\label{fig:disturbance_1}
	\end{figure*}

    For evaluating GANs, we used two types of distributional metrics on the PointNet representation: Fréchet distance~\cite{Shu_2019_ICCV} (named FPD for point clouds), squared maximum mean discrepancy (squared MMD)~\cite{binkowski2018demystifying}.
    This section aims to verify if the metrics can be used for evaluating LiDAR point clouds, since the metrics have been designed for other domains.
    For instance, FPD~\cite{Shu_2019_ICCV} has been proposed for evaluating ShepeNet~\cite{shapenet2015} generation task where each sample forms small-scale point clouds uniformly sampled from CAD objects.
    Squared MMD~\cite{binkowski2018demystifying} was used to extend Fréchet Inception distance (FID)~\cite{NIPS2017_8a1d6947} that is the standard metrics for an image generation task.
    In the image domain, the metrics are known as Kernel Inception distance (KID) in tribute to the Inception feature extractor.
    For the backbone of the feature extractor, we used the off-the-shelf PointNet~\cite{qi2017pointnet} provided by Shu \textit{et al.}~\cite{Shu_2019_ICCV}.
    The PointNet backbone\footnote{\url{https://github.com/seowok/TreeGAN}} is pre-trained on the ShapeNet dataset and used by the original FPD~\cite{Shu_2019_ICCV}.
    To verify if the score is derived from learned features or architecture bias, we compute the metrics using two PointNet encoders with pre-trained weights and random weights.
    All metrics are computed between clean and disturbed sets of KITTI point clouds.
    In Fig.~\ref{fig:disturbance_1}, we provide the results under six types of disturbances; (a) additive Gaussian noises, (b) drop-in Gaussian noises, (c) inflating coordinates, (d) yaw rotation, and (e,f) translation in $x/y$ directions.
    From the results, we can see that both metrics reflect the distributional error if using the pre-trained PointNet.
    We can also see that the metrics sensitive to the translation changes in Fig.~\ref{fig:disturbance_1}c--f.
    Although there are scale gaps depending on the type of disturbance, the results are roughly similar to the sanity check of FID~\cite{NIPS2017_8a1d6947}.
    Therefore, we concluded that the two metrics can be used to evaluate the generative models on LiDAR point clouds.
		
	\section{Generated examples}
	\label{sec:generated_examples}
		
	\begin{figure*}[t]
		\centering
		\includegraphics[width=\hsize]{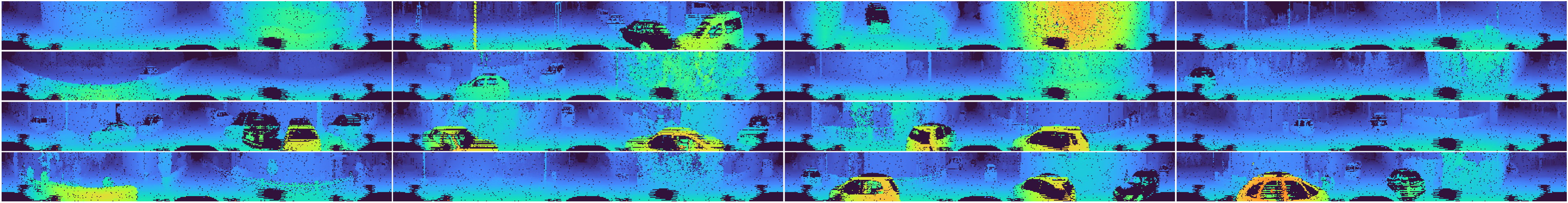}
		\includegraphics[width=\hsize]{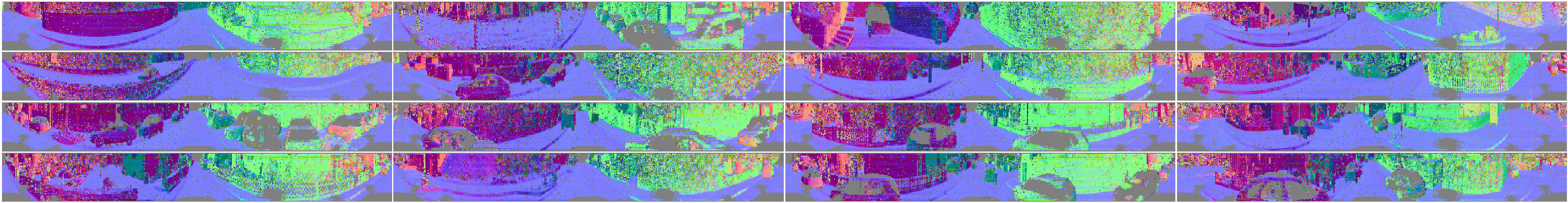}
		Real data
		\\ \vspace{3mm}
		\includegraphics[width=\hsize]{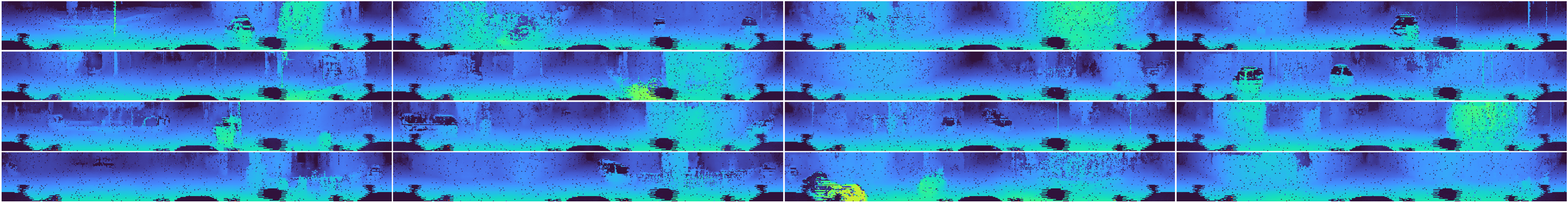}
		\includegraphics[width=\hsize]{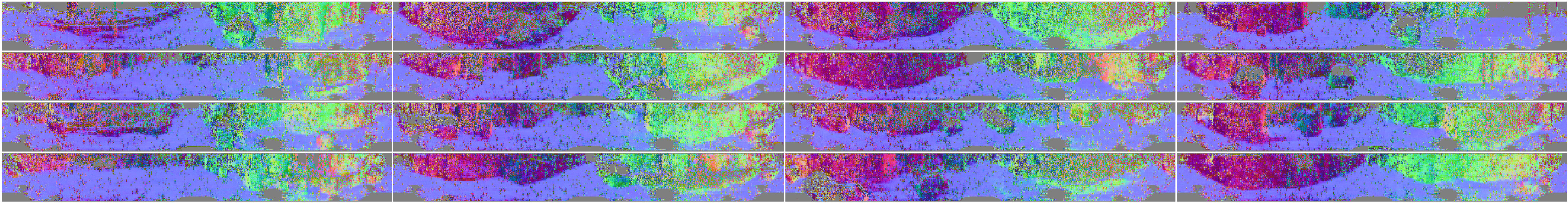}
		Vanilla GAN~\cite{caccia2019deep}
		\\ \vspace{3mm}
		\includegraphics[width=\hsize]{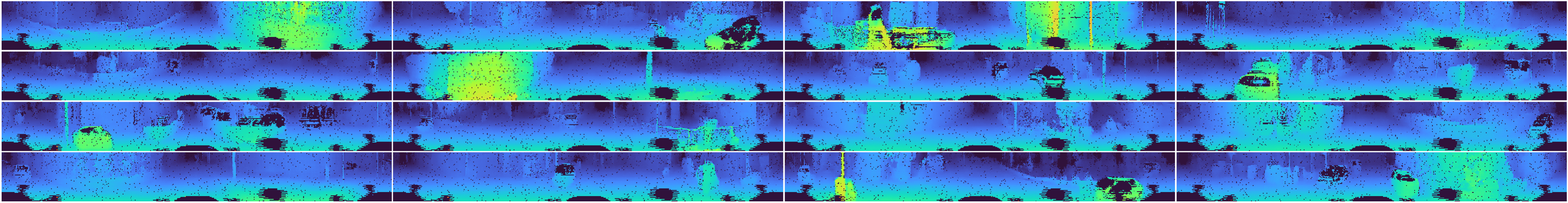}
		\includegraphics[width=\hsize]{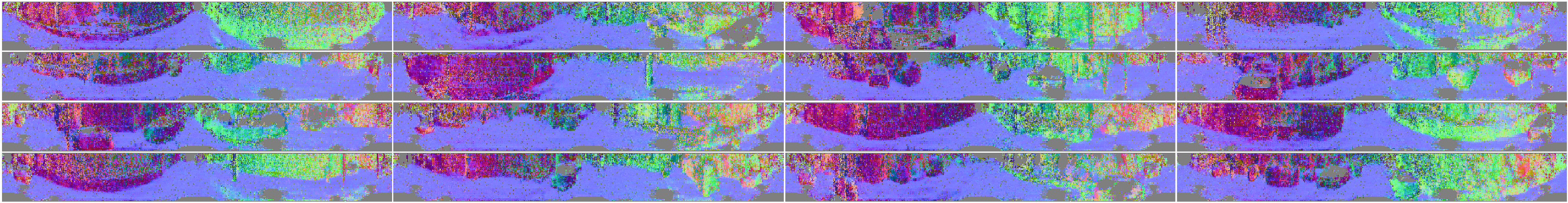}
		DUSty~\cite{nakashima2021learning}
		\\ \vspace{3mm}
		\includegraphics[width=\hsize]{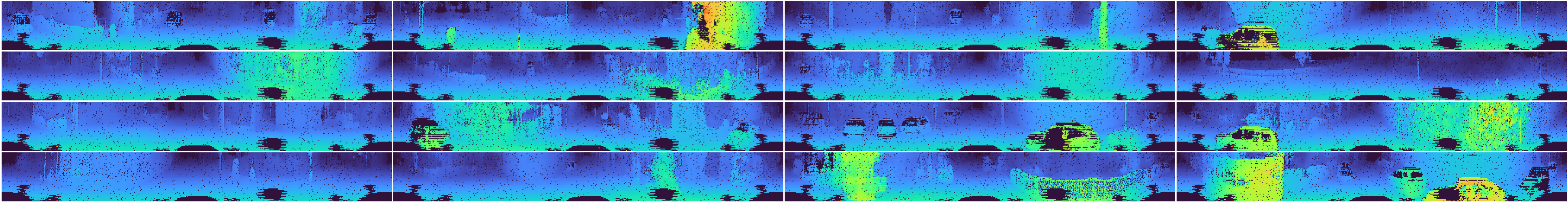}
		\includegraphics[width=\hsize]{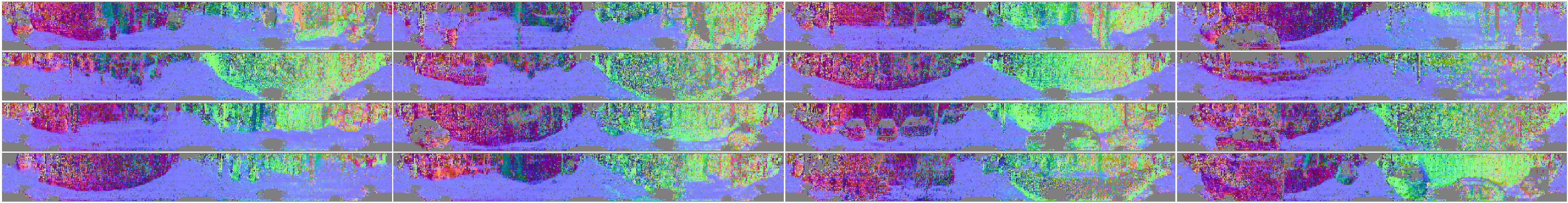}
		\textbf{Ours}
		\caption{Qualitative comparison of uncurated sets of generated samples in the image format (top) and the corresponding surface normal maps (bottom). The surface normal maps are computed from projected Cartesian points.}
		\label{fig:uncurated_depth}
	\end{figure*}

	\begin{figure*}[t]
		\centering
		\includegraphics[width=0.8\hsize]{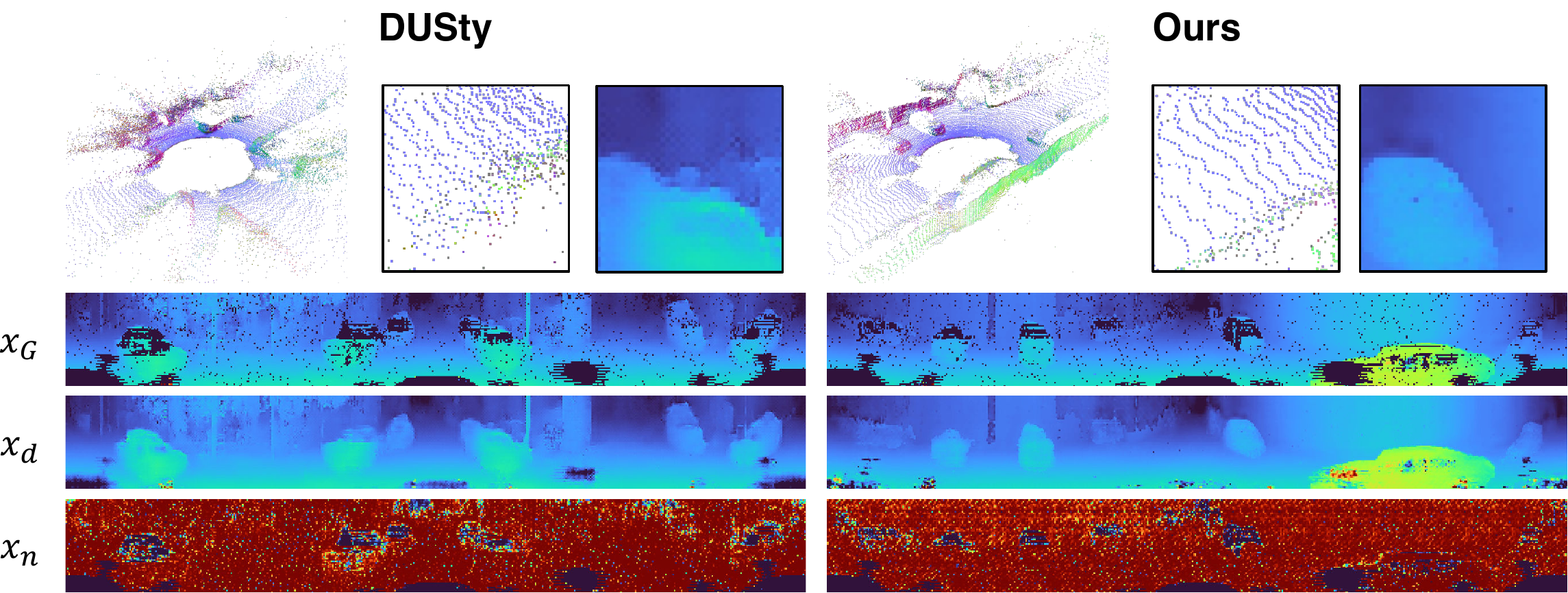}
		\caption{Qualitative comparison between DUSty~\cite{nakashima2021learning} and ours. From top to bottom: generated point clouds, the final inverse depth maps $x_G$, the complete depth maps $x_d$, and the ray-drop probability maps $x_n$.}
		\label{fig:comparison_image_based}
	\end{figure*}
		
	\begin{figure*}[t]
		\centering
		\includegraphics[width=0.6\hsize]{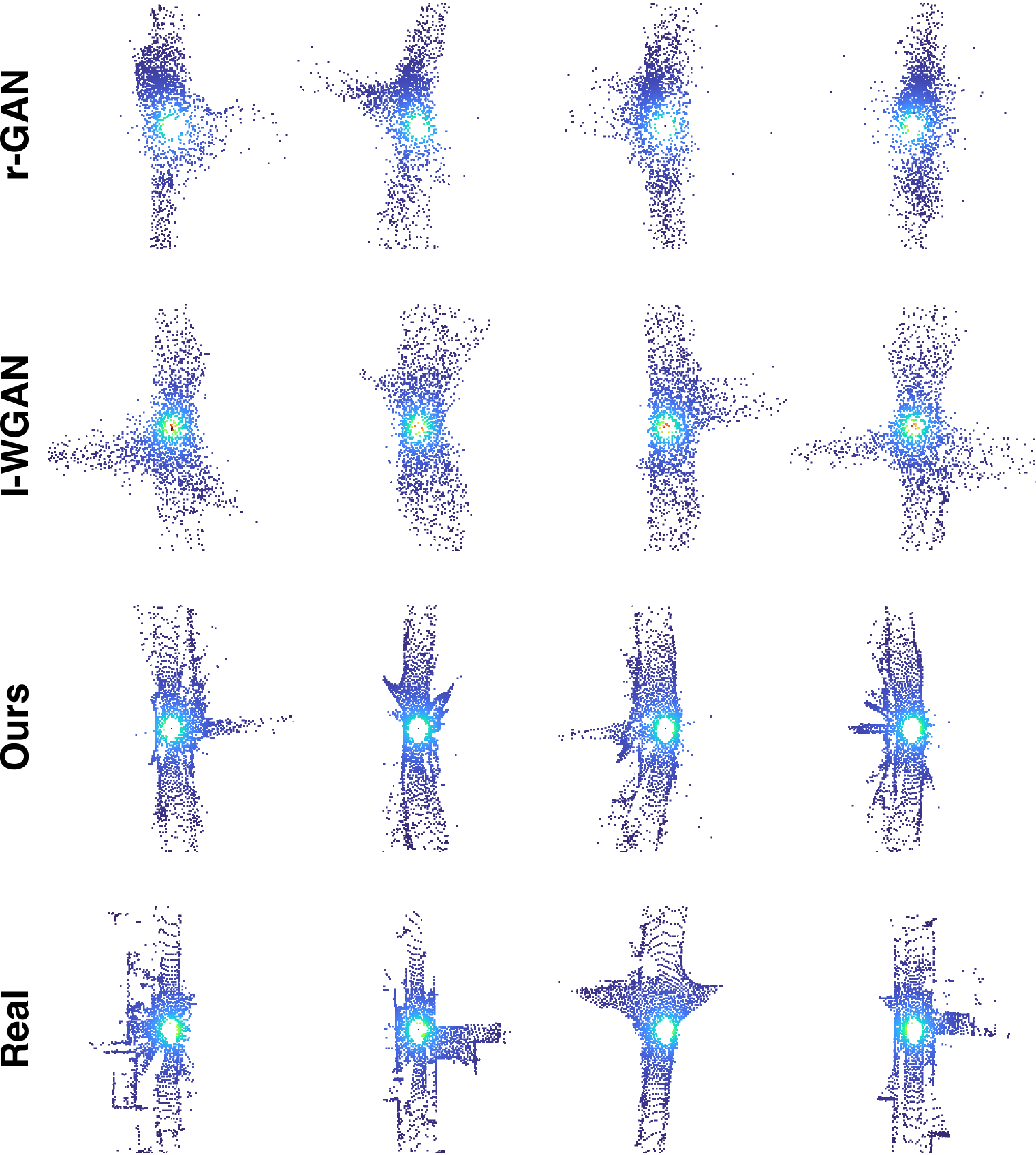}
		\caption{Qualitative comparison in bird's eye views of real and generated point clouds.}
		\label{fig:comparison_point_based}
	\end{figure*}

\begin{figure*}[t]
		\centering
		\setlength\tabcolsep{3pt}
		\begin{tabularx}{\hsize}{CCCC}
			\includegraphics[width=\hsize]{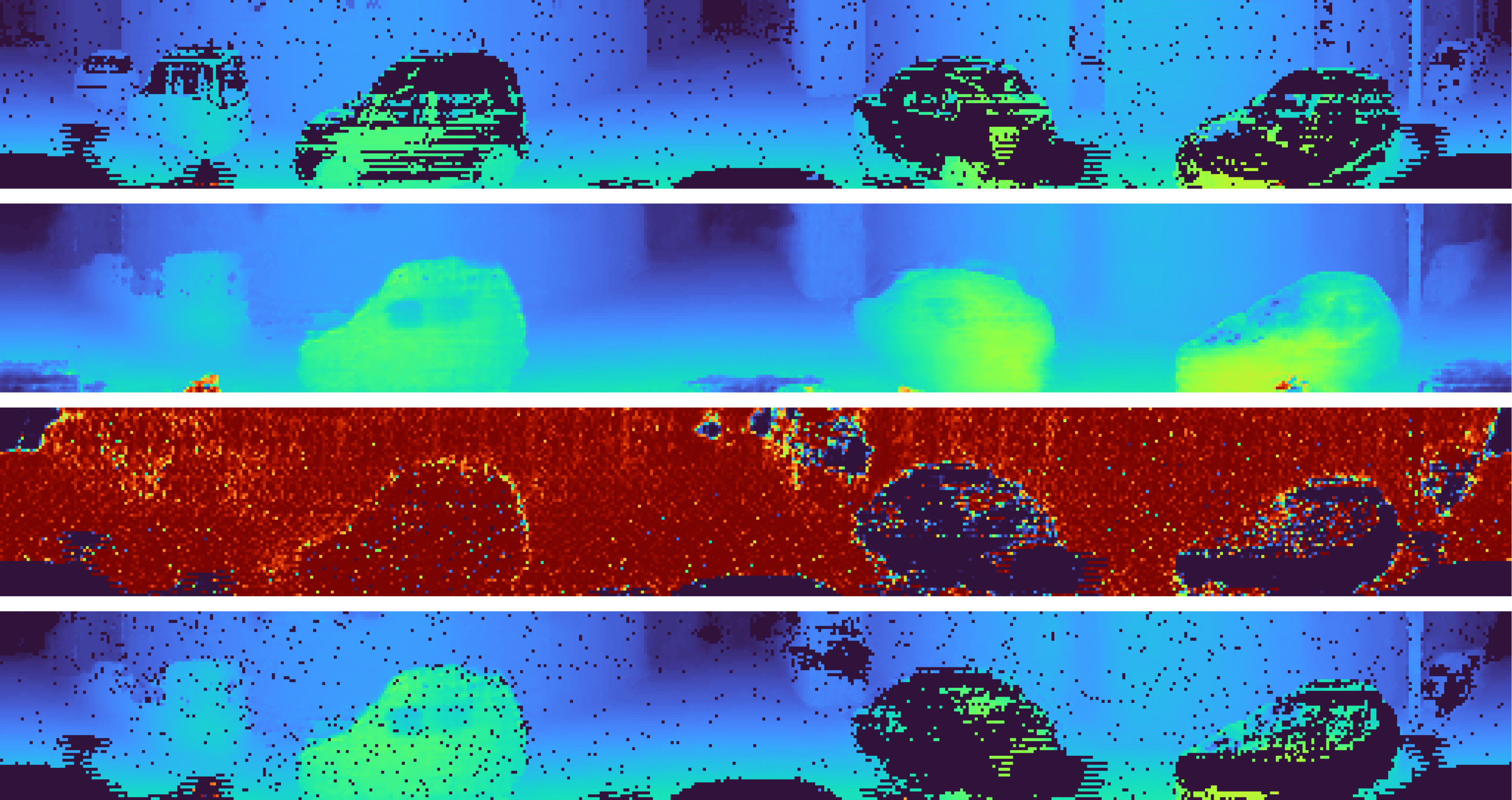}
			  &   
			\includegraphics[width=\hsize]{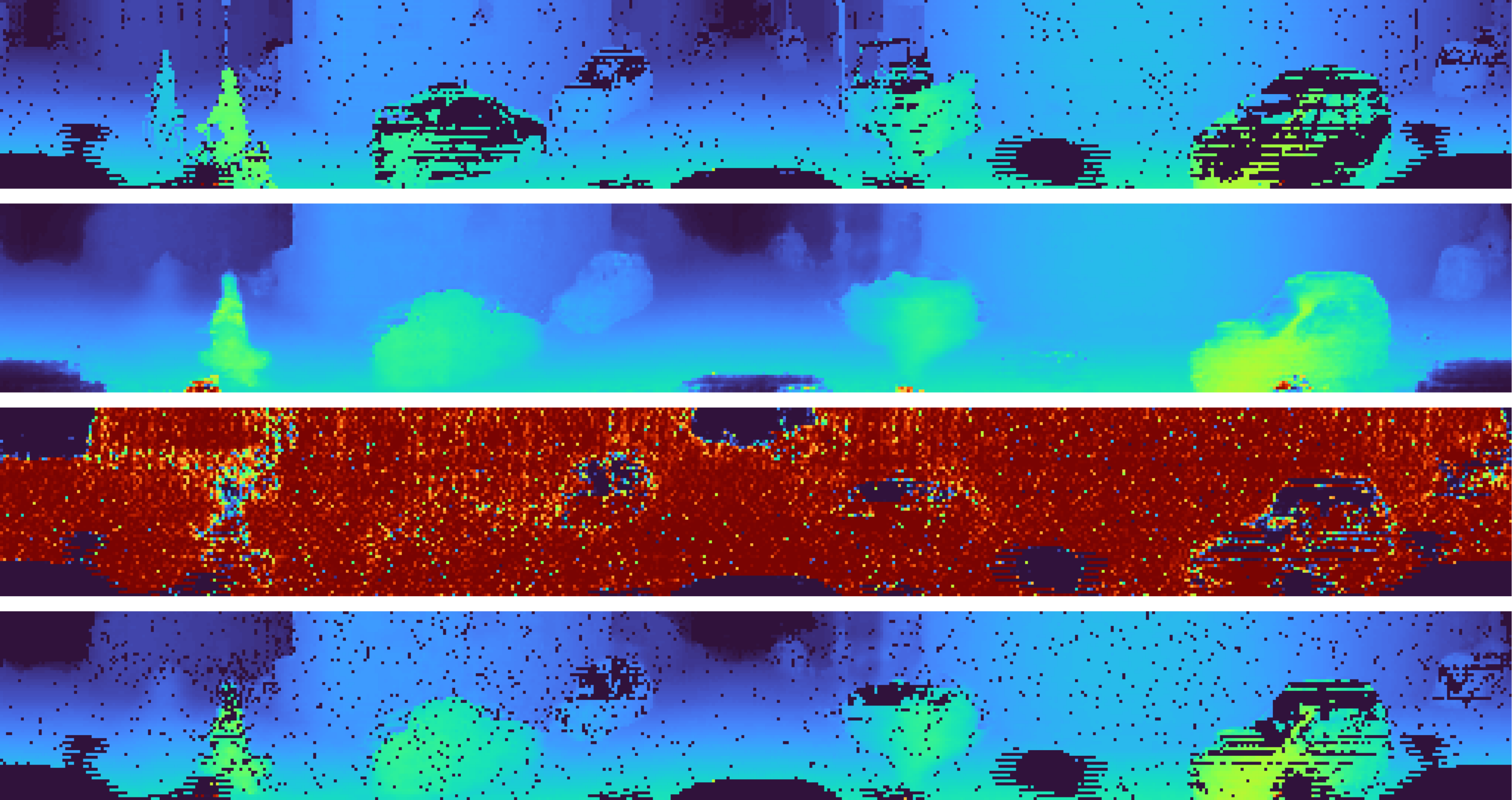}
			  &   
			\includegraphics[width=\hsize]{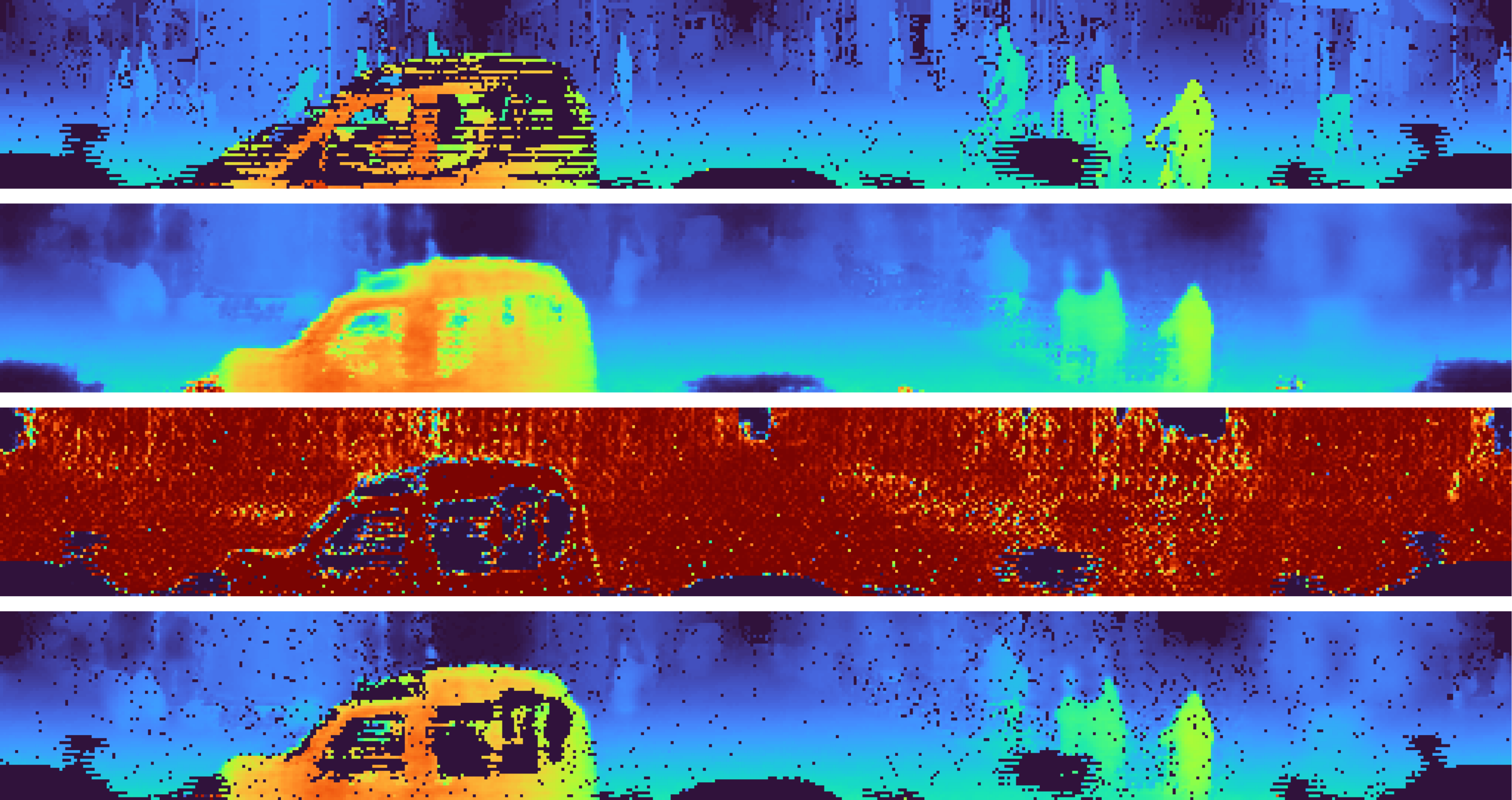}
			  &   
			\includegraphics[width=\hsize]{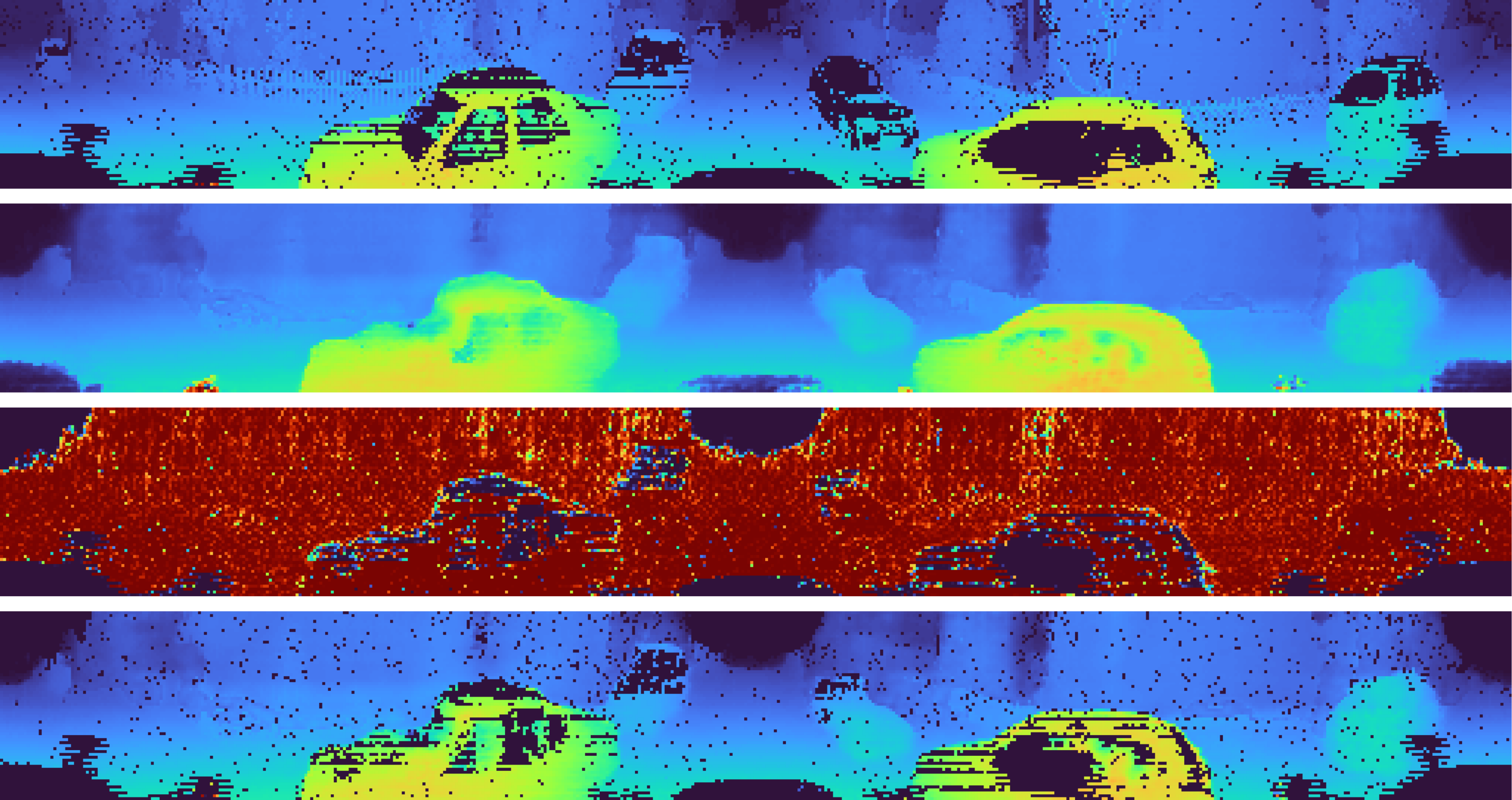}
			\\
			\\
			\includegraphics[width=\hsize]{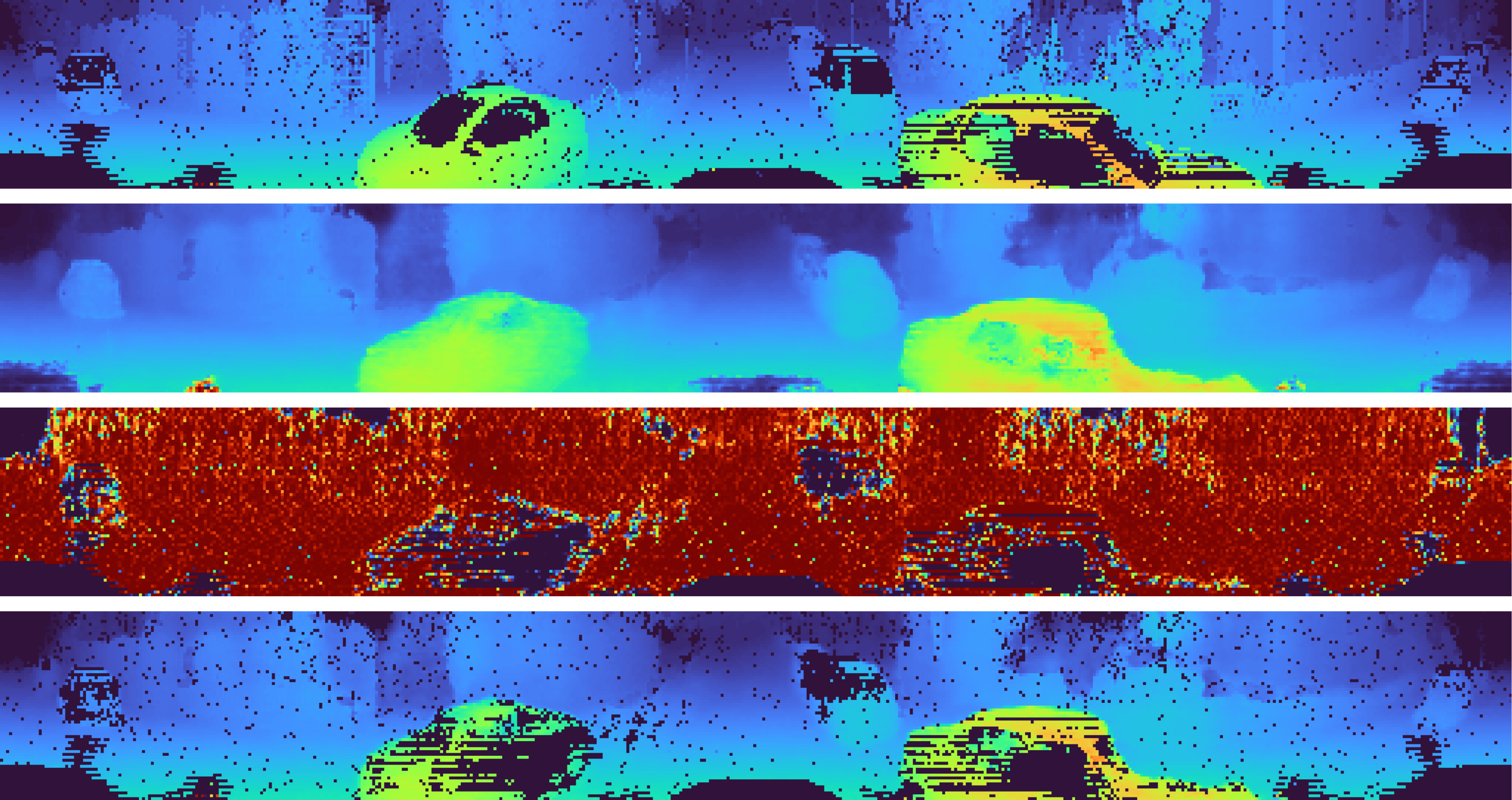}
			  &   
			\includegraphics[width=\hsize]{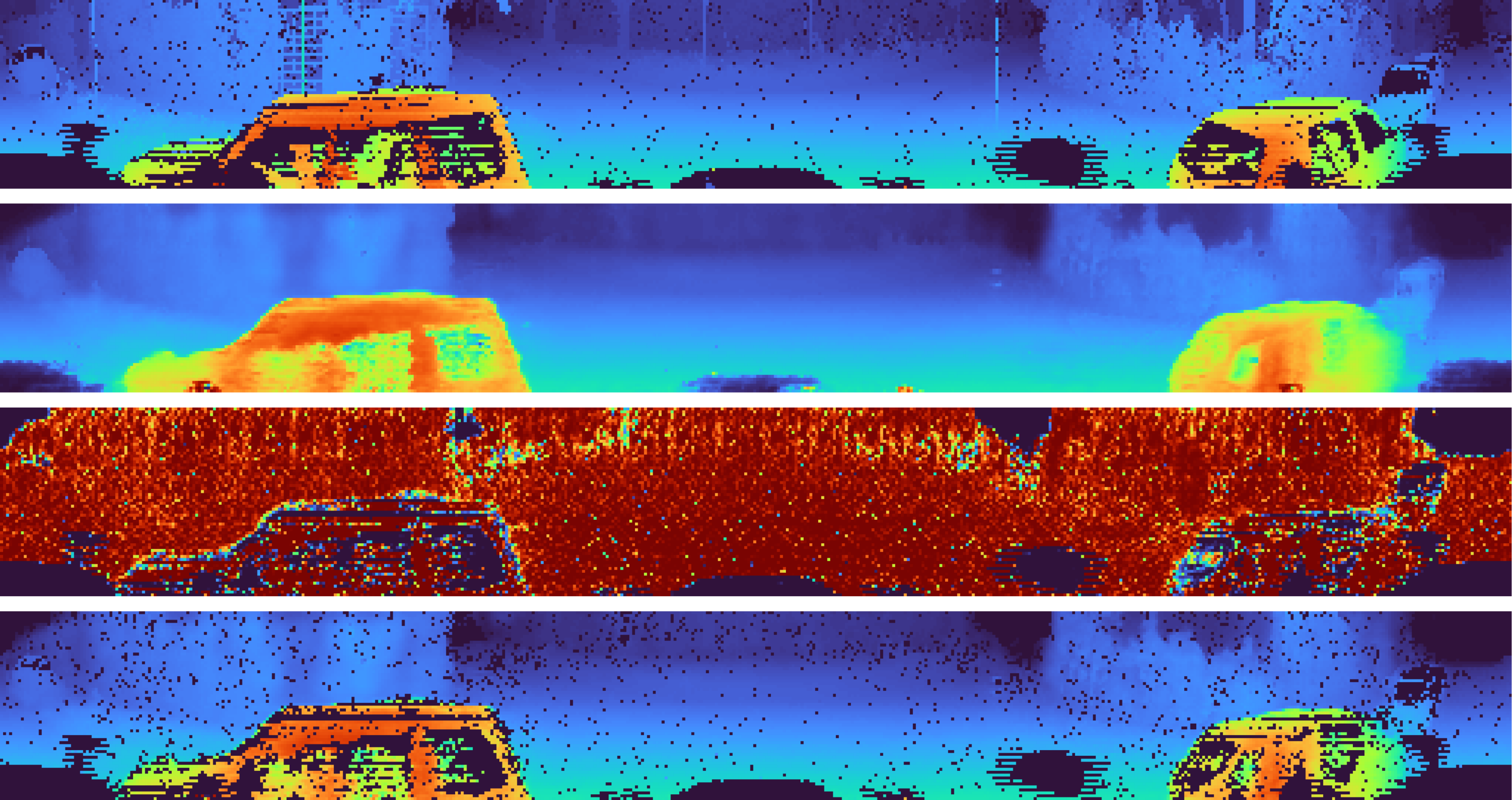}
			  &   
			\includegraphics[width=\hsize]{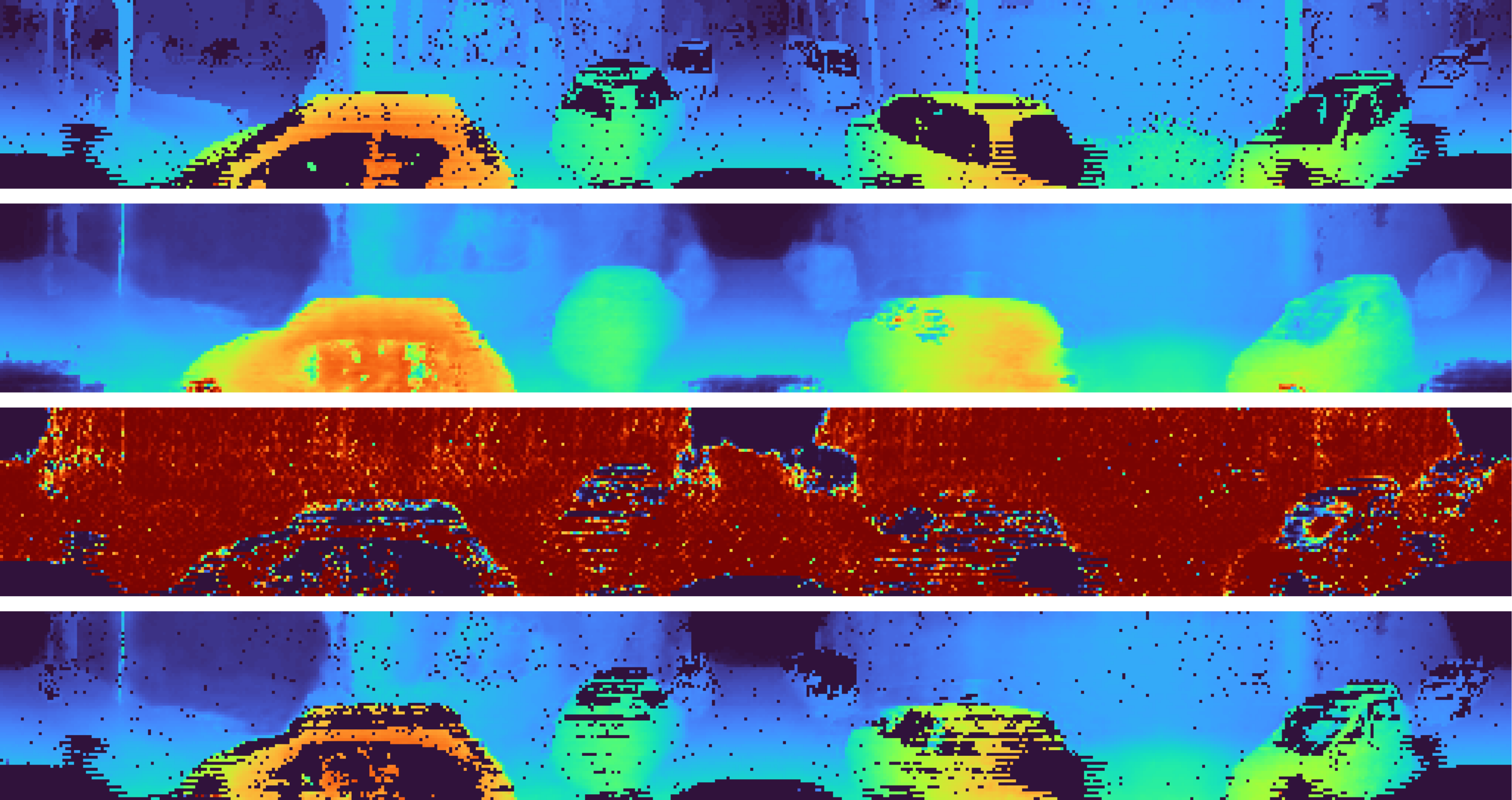}
			  &   
			\includegraphics[width=\hsize]{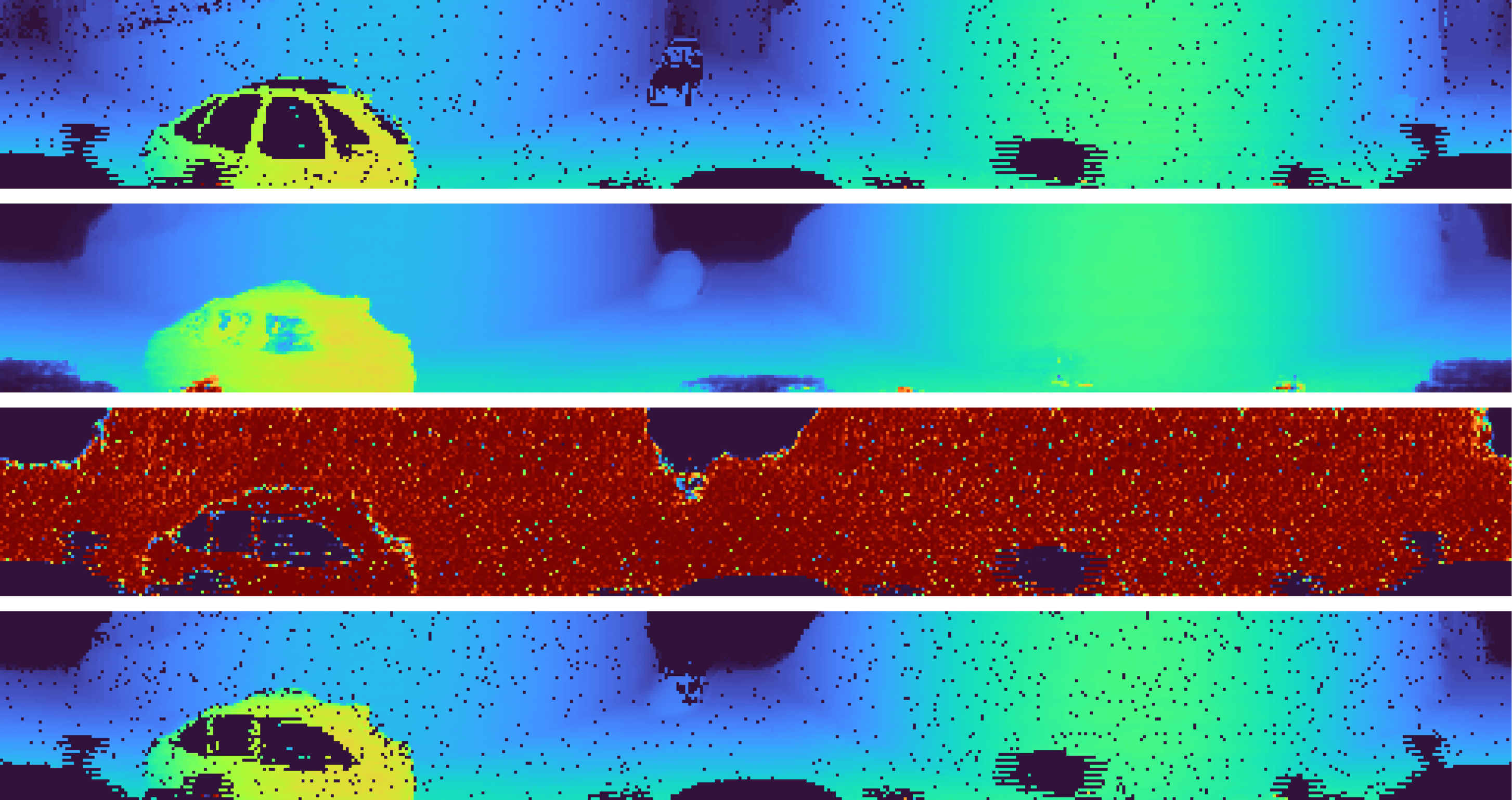}
			\\
		\end{tabularx}
		\caption{Reconstruction examples by our auto-decoding method. For each group, from top to bottom, we show the target range image $\hat{x}$ from KITTI~\cite{geiger2013vision}, the generated inverse depth map $x_d$, the generated ray-drop probability map $x_n$, and the final output $x_G$.}
		\label{fig:autodecoding_examples}
	\end{figure*}

	In Fig.~\ref{fig:uncurated_depth}, we provide uncurated sets of real and generated samples from image-based methods including ours.
	Fig.~\ref{fig:comparison_image_based} compares the results between ours and the most closely related work, DUSty~\cite{nakashima2021learning}.
	A close-up comparison shows that baseline methods include checkerboard artifacts and our method succeeded in expressing the smooth road surface.
	In Fig.~\ref{fig:comparison_point_based}, we provide uncurated sets of real and generated samples from point-based methods and ours.
	Our method is superior in point density distribution and edges.
	In Fig.~\ref{fig:autodecoding_examples}, we show reconstruction examples by our auto-decoding method.
	From the real data via the lossy measurement, our model produced the smooth shapes and the reasonable ray-drop probability maps.
	For instance, the ray-drop probabilities have uncertainty on the object edges.
		
	\section{Sim2Real semantic segmentation}
	\label{sec:sim2real_examples}
		
\begin{figure*}[t]
		\footnotesize
		\begin{tabularx}{\hsize}{CCCCC}
			Input depth & Ground truth & Config-A & Config-C~\cite{wu2019squeezesegv2} & Config-E (\textbf{ours}) 
		\end{tabularx}
		\\
		\includegraphics[width=\hsize]{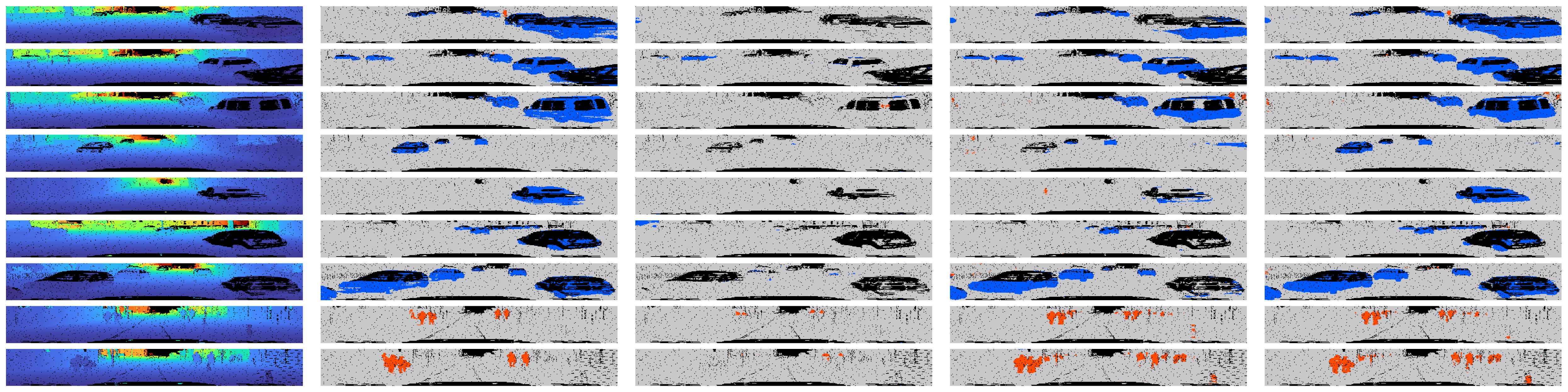}
		\\
		\begin{tabularx}{\hsize}{CCCCC}
			\includegraphics[height=4mm]{figs/colorbars/depth.pdf}                   &   
			\includegraphics[height=4mm]{figs/colorbars/labels.pdf} &   &   &   
		\end{tabularx}
		\caption{Comparison of Sim2Real segmentation results in 2D range images. We compare three types of ray-drop priors from our main paper. Config-A: GTA-LiDAR without ray-drop rendering. Config-C: GTA-LiDAR with Bernoulli noises from the pixel-wise frequency~\cite{wu2018squeezeseg}. Config-E (ours): GTA-LiDAR with Bernoulli noises from our auto-decoded ray-drop probability.}
		\label{fig:segmentation}
	\end{figure*}

	In Fig.~\ref{fig:segmentation}, we show Sim2Real segmentation results on KITTI-frontal~\cite{wu2019squeezesegv2}.
	All models are trained on GTA-LiDAR while the ray-drop priors are different.
	We can see that our method (config-E) greatly improved the false negative regions of car classes.
		
\end{appendices}

\end{document}